%% file: main.tex
\ifdefined\XeTeXrevision\else\ifdefined\pdfoutput\pdfoutput=1 \fi\fi
\documentclass{article}
\usepackage[preprint]{neurips_2026}
\usepackage[utf8]{inputenc}
\usepackage[T1]{fontenc}
\usepackage{hyperref}
\usepackage{url}
\usepackage{booktabs}
\usepackage{amsfonts}
\usepackage{nicefrac}
\usepackage{xcolor}
\usepackage{colortbl}
\usepackage{makecell}
\usepackage{graphicx}
\usepackage{multirow}
\usepackage{amsmath}
\usepackage{amssymb}
\usepackage{array}
\usepackage{tikz}
\usetikzlibrary{positioning,arrows.meta,fit,calc,backgrounds}

\graphicspath{{figs/}}

\newcommand{\cmk}{\ensuremath{\checkmark}}

\usepackage{algorithm}
\usepackage{algpseudocode}

\title{Physics-R1: An Audited Olympiad Corpus\\and Released Verifiers for Visual Physics Reasoning}

\author{%
  Shan Yang\\
  Independent Researcher\\
  \texttt{alexyangshan@gmail.com}
}

\begin{document}

\maketitle

\begin{abstract}
Trackable improvement in multimodal physics reasoning rests on a training-and-evaluation system that is itself rarely verified: the corpora a model trains on, the reward it is optimized against, and the benchmarks and judges that score it. We audit this system end to end and find that standard construction practices systematically distort measurement: contamination slips past $n$-gram deduplication, translation degrades problems, saturated multiple-choice formats overstate capability, partial-credit training rewards are easier to exploit than to earn, and open-ended grading silently depends on the choice of judge. Left unverified, these distortions inflate reported progress and leak test knowledge into training. We answer with a released verifier system: a three-stage contamination audit that certifies the train/test boundary behind an audited multimodal training corpus and a held-out olympiad benchmark; a binary answer verifier that supplies the reinforcement-learning training reward; and an answer-judging harness that brackets every open-ended score between a deterministic strict layer and a large-language-model liberal layer. All per-record verdicts are released and cross-checked against an independent open-weight judge, whose substitution shifts absolute scores but preserves the sign of every base-to-trained lift. Training against the binary answer verifier confirms the certified corpus supports training: a reference recipe lifts an 8B open-source base by $18.3$ points on the held-out benchmark across three seeds. The simple binary reward also beats a dense partial-credit variant on three of four open-ended benchmarks, tying the fourth. All verifiers, verdicts, and datasets are public.
\end{abstract}

% =====================================================================
% Body sections — split into sections/<section>.tex
% =====================================================================

\input{sections/intro.tex}
\input{sections/related.tex}
\input{sections/data.tex}
\input{sections/method.tex}
\input{sections/results.tex}
\input{sections/analysis.tex}

\section*{Acknowledgments}
The author thanks Kevin Zhou for granting permission to redistribute his olympiad handouts under CC BY-NC 4.0, the Estonian Physics Olympiad committee for making their archived problems and solutions publicly available at \url{https://fyysika.ee/}, the international olympiad committees (IPhO, NBPhO, EuPhO, APhO, USAPhO, INPhO) for the public archives that enabled the novel-source held-out evaluation, and the maintainers of the public physics-VL benchmarks (PhyX, MMMU-Pro, OlympiadBench, UGPhysics, PhysReason, PhysUniBench) whose released training pools and evals enabled the contamination audit reported in this work. Compute support for Physics-R1 training and evaluation was provided by RunPod.

\bibliographystyle{plainnat}
\bibliography{references}

\appendix

% =====================================================================
% Appendix sections — split into appendix/<section>.tex
% =====================================================================

\input{appendix/audit.tex}
\input{appendix/splits.tex}
\input{appendix/reward.tex}
\input{appendix/judge.tex}
\input{appendix/license.tex}
\input{appendix/repro.tex}
\input{appendix/datasheet.tex}
\input{appendix/extras.tex}

\end{document}

%% file: sections/intro.tex
% =====================================================================
\section{Introduction}\label{sec:intro}
% =====================================================================

Physics olympiad problems are among the most demanding tests of multimodal scientific reasoning: solving one means reading a diagram, selecting the governing physical law, and carrying symbolic and numerical work through to a checkable final answer. As vision--language models (VLMs) post rapidly rising scores on physics benchmarks, those scores increasingly stand in for claims about scientific reasoning ability and steer what gets trained next. The challenge is that reliably improving a VLM's physics reasoning rests on an entire verifiable training-and-evaluation system: the training data, the training reward verifier, and a valid benchmark with a fixed judge. Each component is a place where measurement can silently break, yet only the final score is usually inspected. Five common practices undermine the system from within. Training pools aggregate public problem collections and are deduplicated against evaluations by a single n-gram pass; paraphrases survive it, so a model may be scored on lightly reworded versions of its own training data, and recall reads as reasoning. Benchmarks are evaluated in English even when their problems were composed in another language, as if translation were free; wording, idiom, and notation shift in translation, and the score shifts with them. Multiple-choice question (MCQ) formats compress scores against the frontier ceiling, where guessing floors and option elimination do part of the work, so differences in genuine capability stop registering. In reinforcement-learning post-training, rewards shaped with partial credit pay out for answer form as well as physics, and a policy under optimization pressure collects the form. And when answers are open-ended, the grade depends on the judge itself: which model grades, under what prompt, at what leniency. That instrument is almost never held fixed or released. We therefore treat the system itself as the object to be verified: this paper audits every component, states each failure as a numbered finding, and releases the verifiers and artifacts constructed to close each gap (Figure~\ref{fig:pipeline}).

Auditing every component yields five findings, from the training data to the evaluation judge. \textbf{Finding 1 (contamination):} a standard single-stage 5-gram Jaccard audit passes public physics training pools as clean against six public evaluations. Our three-stage audit of n-gram overlap, then embedding cosine, then a large language model (LLM) judge surfaces 134 near-duplicates of evaluation problems in SciInstruct alone (\S\ref{sec:data:audit}). \textbf{Finding 2 (translation):} on 59 paired Estonian/English versions of identical olympiad problems, the same frontier model scores 17 percentage points (pp) higher on the Estonian originals (\S\ref{sec:results:cross}). \textbf{Finding 3 (format saturation):} identical model weights score 46 points lower on our novel-source open-ended olympiad evaluation than on a public MCQ benchmark, a spread produced by evaluation format, problem genre, and source novelty on fixed weights (\S\ref{sec:results:dataset}). \textbf{Finding 4 (reward robustness):} training against the binary answer verifier beats a dense partial-credit variant on three of four open-ended benchmarks and ties the fourth, by up to $8.9$\,pp at matched training steps; under optimization pressure, shaped partial credit rewards format proxies rather than physics (\S\ref{sec:results:ablations}). \textbf{Finding 5 (judge reliability):} re-scoring every re-derivable open-ended column with a local open-weight judge shifts absolute scores by up to $+37$\,pp and can reverse cross-model rankings, so comparisons are meaningful only under a fixed, released judge (\S\ref{sec:analysis}).

The release is organized as a single verifiable system (Figure~\ref{fig:pipeline}): a certified data boundary, training that optimizes against a verifier, and measurement through a released judging harness. Three deployed verifiers anchor it. Verifier~1 is the three-stage contamination audit: it certifies a train/test boundary between any training pool and any evaluation, and ships as runnable code with per-pair verdicts (\S\ref{sec:data}). Its certified artifacts are \textsc{PhysCorp-A}, a $6{,}432$-record audited multimodal corpus; \textsc{PhysR1Corp}, a $2{,}268$-record closed-form pool for reinforcement learning (RL); and \textsc{PhysOlym-A}, a $500$-problem held-out open-ended olympiad benchmark with organizer difficulty labels and a bilingual English/Estonian subset (\S\ref{sec:data:physolym}). Verifier~2 is the binary answer verifier whose verdict is the RL training reward (\S\ref{sec:method}). Verifier~3 is the answer-judging harness that scores open-ended physics answers: a deterministic strict layer, plus an LLM liberal layer whose per-record verdicts are released (\S\ref{sec:eval:judge}).

The final artifact, Physics-R1, is a reference RL recipe, included as evidence that the certified corpus supports training rather than as a method contribution; it closes the loop on all three verifiers, training inside the boundary Verifier~1 certifies, optimizing against Verifier~2, and scored by Verifier~3. Cold-started from an open-source 8B VLM with a binary correctness reward, it lifts the base by $18.3$\,pp on \textsc{PhysOlym-A} at the 3-seed mean (\S\ref{sec:results}). Only the strongest frontier model remains ahead of it, by $7.1$\,pp.

\input{figures/fig1_pipeline}

% =====================================================================

%% file: figures/fig1_pipeline.tex
\begin{figure}[t]
\centering
\includegraphics[width=\linewidth,trim=33 79 26 26,clip]{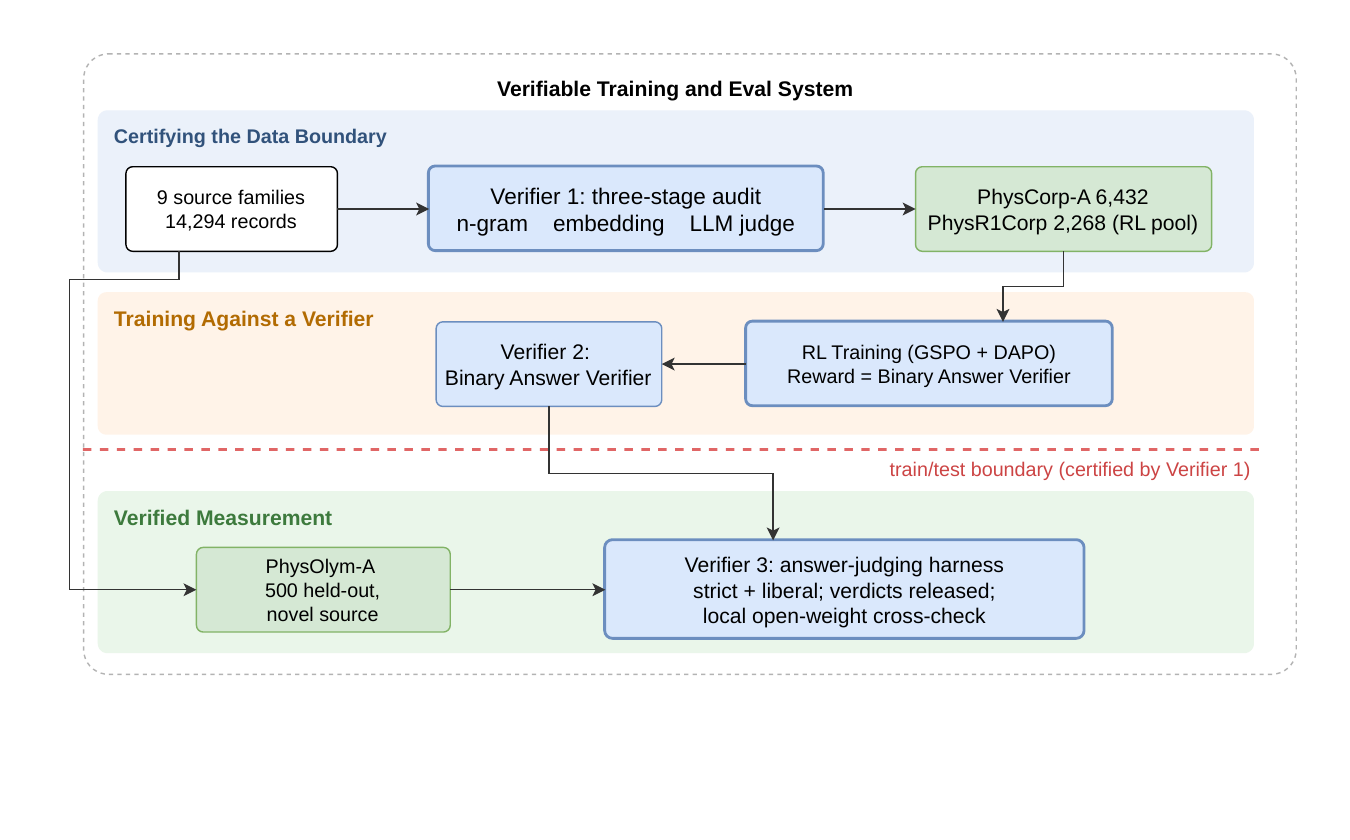}
\caption{\textbf{The verifiable training-and-evaluation system overview.} Verifier~1, the three-stage audit, certifies the corpora and the held-out benchmark on opposite sides of the train/test boundary. Training optimizes against Verifier~2, the binary answer verifier, a deterministic oracle of the same family as Verifier~3's strict layer. All reported numbers flow through Verifier~3, the answer-judging harness, whose per-record verdicts are released and cross-checked by a local open-weight judge (Appendix~\ref{app:judge}).}
\label{fig:pipeline}
\end{figure}

%% file: sections/related.tex
% =====================================================================
\section{Related Work}\label{sec:related}
% =====================================================================

\paragraph{Rule-based RL for reasoning.} DeepSeek-R1~\citep{guo2025deepseekr1} established that simple rule-based rewards suffice to train competitive math reasoners, via GRPO~\citep{shao2024deepseekmath}; MM-Eureka~\citep{meng2025mmeureka} extended the recipe to VLMs with a difficulty curriculum, DAPO~\citep{yu2025dapo} added decoupled clipping and dynamic sampling, and GSPO~\citep{zheng2025gspo} moved to sequence-level importance weighting. Physics-R1 inherits these components unchanged. Although physics steps carry units and conservation laws that admit per-step verification, a binary reward is the better default under group-normalized advantages (\S\ref{sec:method:theory}); the dense physics-native reward is an ablation.

\paragraph{Physics QA benchmarks.} PhyX~\citep{shen2025phyx}, OlympiadBench-Physics~\citep{he2024olympiadbench}, UGPhysics~\citep{xu2025ugphysics}, PhysReason~\citep{zhang2024physreason}, MMMU/MMMU-Pro~\citep{yue2024mmmu,yue2024mmmupro}, MMK12~\citep{meng2025mmeureka}, PHYBench~\citep{qiu2025phybench}, and PhysUniBench~\citep{wang2025physunibench} are the canonical references. Only PHYBench, OIBench, and PutnamBench publish a contamination protocol. None publish the three-stage pairwise audit of n-gram overlap, embedding cosine, and LLM judgment that we introduce in \S\ref{sec:data:audit}. Table~\ref{tab:benchmark_comparison} (Appendix~\ref{app:extras:tables}) maps our released artifacts against related benchmarks on nine axes.

\paragraph{Contamination audits and other prior work.} PutnamBench~\citep{tsoukalas2024putnambench}, FrontierMath~\citep{glazer2024frontiermath}, HLE~\citep{phan2025hle}, and EnigmaEval~\citep{wang2025enigmaeval} provide release-policy templates and dismissal grounds; methodological work spans n-gram audits~\citep{sainz2023nlp}, the rephrased-samples failure mode~\citep{yang2023rephrased} (which our Stage~2 catches), contamination-impact measurement~\citep{singh2024beyond}, and performance-based detection~\citep{dekoninck2024constat}; the survey of \citet{ravaut2024constam} consolidates these. We import the math template, adding the embedding-cosine pass because physics statements (units, vectors, figure references) are more paraphrase-sensitive than typical math problems---a sensitivity Table~\ref{tab:audit_thresholds} quantifies. PhysBench~\citep{chow2025physbench} evaluates physical-world understanding over interleaved video-image-text tasks, an orthogonal scope. Multilingual benchmarks have proliferated~\citep{xuan2025mmlluprox,wang2024multilingualcontam,wu2025bitter}; our cross-lingual finding (\S\ref{sec:results:cross}) instead evaluates 59 identical problems in original Estonian and in English translation on the same model with paired tests, isolating a within-problem effect.

% =====================================================================

%% file: sections/data.tex
% =====================================================================
\section{The Verifiable System I: Certifying the Data Boundary}\label{sec:data}
% =====================================================================

The system's first stage certifies its data boundary (Figure~\ref{fig:pipeline}). Verifier~1, the released three-stage contamination audit, performs the certification: its input is any pair of problem pools; its output is a verdict for every candidate overlap, plus the certified-disjoint pools. This section describes the verifier's construction and measured failure modes (\S\ref{sec:data:audit}), then the artifacts it certifies: \textsc{PhysCorp-A} (\S\ref{sec:data:composition}), \textsc{PhysR1Corp}, and the held-out \textsc{PhysOlym-A} (\S\ref{sec:data:physolym}). All artifacts carry per-source licenses and Croissant metadata (Appendix~\ref{app:license}); release and replay details are in Appendix~\ref{app:repro}.

\subsection{Construction and Failure Modes}\label{sec:data:audit}

\input{tables/table4_audit}

The verifier applies one definition of contamination pairwise between a training pool, four external corpora, and the held-out splits. \emph{Stage~1 (n-gram recall):} tokenize each problem statement, build its 5-gram shingle set, and flag pairs with Jaccard overlap $J \ge 0.4$. \emph{Stage~2 (paraphrase recall):} embed each statement with \texttt{mxbai-embed-large} and flag pairs with cosine similarity $\ge 0.85$; this catches the reworded duplicates Stage~1 misses, but also flags problems that merely share a topic. \emph{Stage~3 (precision filter):} an LLM judge (Haiku 4.5) reads both statements of each Stage-2 candidate and classifies the pair as a \emph{close duplicate} (a paraphrase or numeric variation of the same problem) or a \emph{same-topic neighbor} (related physics, distinct setup). Only close duplicates are removed. Pseudocode is in Algorithm~\ref{alg:audit} and worked examples in Appendix~\ref{app:audit}.

Table~\ref{tab:audit} shows why all three stages are needed (Finding~1). The three public training pools we re-audit all pass Stage~1 with zero hits against all six public evals, yet Stage~2 surfaces $4{,}846$ close-content candidate pairs in SciInstruct alone. Of these, Stage~3 confirms $134$ as near-duplicates of evaluation problems. The close-duplicate rate is sharply cosine-driven ($100\%$ at $\cos\ge 0.95$ down to $1.5\%$ at the threshold edge; Appendix~\ref{app:audit}, Table~\ref{tab:audit_haiku}): cosine provides the recall, the LLM judge the precision. \textsc{PhysCorp-A} retains zero Stage-2 candidates by construction; \textsc{PhysR1Corp} drops $165$ near-duplicates (Appendix~\ref{app:audit:leakage}), and its $19$ remaining Stage-2 candidates are all same-topic neighbors.

\paragraph{Measured failure modes of Verifier~1.}\label{sec:data:leakage}
A verifier is only trustworthy if its own failure modes are measured, so we re-audited a $1{,}679$-record sample of the released pre-audit pool, a sample curated under the field-default workflow of 5-gram deduplication. Against an internal 500-record evaluation, the single-stage audit detects a $3.3\%$ leak rate, all exact copies. Adding the embedding stage raises the detected rate to $8.8\%$; the gap is paraphrased leakage that no n-gram check can see. The detected rate is also threshold-sensitive, sweeping $4.7$--$27.1\%$ as the cosine cutoff moves from $0.90$ to $0.80$ (Appendix~\ref{app:audit:leakage}, Table~\ref{tab:audit_thresholds}), which is why released verdicts always ship with their operating thresholds. Stage-3 precision rests on the LLM judge plus manual spot checks; a human-labeled validation on a cosine-stratified sample is left to follow-up work.

\subsection{The Certified Corpora}\label{sec:data:composition}

The corpus is drawn from nine source families (Table~\ref{tab:corpus}, Appendix~\ref{app:extras:more-tables}). Five families are repackaged from existing collections under documented licenses: UGPhysics~\citep{xu2025ugphysics}, OpenStax College and University Physics~\citep{openstax2024physics}, Physics Stack Exchange~\citep{stackexchange2024physics}, an MMMU seed with o1 chain-of-thought (CoT) traces~\citep{yue2024mmmu}, and PhysReason~\citep{zhang2024physreason}. Four families contribute material new to machine-readable form. The Estonian Physics Olympiad collection~\citep{efo} contributes 418 problems spanning 2004--2018, each carrying an organizer-issued 1--10 difficulty label. Of these, 201 carry official English translations issued by the olympiad organizers, forming the bilingual English/Estonian (EN/ET) subset; the remaining 217 exist only in Estonian. We deliberately do not machine-translate the Estonian-only problems: that would inject into our own release exactly the confound Finding~2 documents. Kevin Zhou's olympiad handouts~\citep{zhou2018handouts} contribute 692 problems with native 1--5 point values, with inline secondary-source attribution preserved per record (Appendix~\ref{app:license}). Refreshed scrapes of six international olympiads (IPhO, NBPhO, EuPhO, APhO, USAPhO, INPhO) complete the pool~\citep{ipho2025archive,nbpho2025archive,eupho2025archive,apho2025archive,usapho2025archive,inpho2025archive}. Per-source licenses are listed in Appendix~\ref{app:license} and carried through to each record.

\textsc{PhysCorp-pre-audit} ships in full, so downstream users can re-run the verifier from scratch. \textsc{PhysCorp-A} is the $6{,}432$-record subset that survives all three audit stages plus a re-audit against PhysReason-full and PhysUniBench-en; that re-audit drops a further 804 records, mostly against PhysReason-full. The released pool is disjoint, at the joint operating thresholds, from all six public evaluations we audit against, including the held-out \textsc{PhysOlym-A}. The candidate-to-release cleanup for \textsc{PhysR1Corp} is described in \S\ref{sec:data:audit}.

Approximately $73$ \textsc{PhysR1Corp} records ($3.2\%$) have LLM-touched problem statements, documented per-distribution in the Croissant metadata; \textsc{PhysOlym-A} contains no synthetic problem content.

\subsection{\textsc{PhysOlym-A}: The Certified Held-Out Benchmark}\label{sec:data:physolym}

Standard physics-VL benchmarks no longer resolve frontier-class differences. PhyX clusters top entries within ten points of its $80\%$ ceiling; OlympiadBench-Physics predates the contamination-audit discipline; UGPhysics is itself a candidate for audited training data rather than held-out evaluation. Physics-R1's headline evaluation and reward ablation depend on a held-out signal that is non-saturating and contamination-clean against the training pool.

\textsc{PhysOlym-A} (Physics Olympiad, Audited) contains 500 problems: 200 from Kevin Zhou's handouts, 136 from the Estonian Physics Olympiad collection, 85 from an IPhO/NBPhO/EuPhO scrape, and 79 from an APhO/USAPhO/INPhO scrape. $499$ of the 500 are novel-source under our four-corpus audit. Native difficulty signals are preserved: 131 of the 136 Estonian records carry organizer-issued 1--10 difficulty labels, and $38\%$ of all records carry Zhou's 1--5 point values. Verifier~1 certifies zero Stage-3 near-duplicate overlaps between the audited training pool and \textsc{PhysOlym-A}. The single non-novel record, an EuPhO 2020 problem shared with OlympiadBench-Physics, is disclosed in Appendix~\ref{app:audit} rather than silently dropped. Scoring is performed by Verifier~3 (\S\ref{sec:eval:judge}).

% =====================================================================

%% file: tables/table4_audit.tex
% Table: train-vs-test contamination between our two released training pools and the public eval splits.
\begin{table}[t]
\centering
\caption[Train/test contamination matrix, three-stage audit]{\textbf{A standard single-stage audit calls every public training pool clean; the three-stage audit surfaces 134 near-duplicates of evaluation problems and certifies both released pools clean.} Rows are public physics eval splits; columns are training pools. Each cell reports Stage-1 / Stage-2 / Stage-3 pair counts under Algorithm~\ref{alg:audit}: Stage-1 is 5-gram Jaccard $\ge 0.4$, Stage-2 is embedding cosine $\ge 0.85$, and Stage-3 is the LLM-judge precision filter separating close duplicates from same-topic neighbors (analysis in \S\ref{sec:data:audit}).}
\label{tab:audit}
\scriptsize
\setlength{\tabcolsep}{2.5pt}
\begin{tabular}{lccc|cc}
\toprule
\multirow{2}{*}{Eval $\downarrow$ \,/\, Train pool $\to$}
& \multicolumn{3}{c|}{\emph{Other published pools (we re-audit)}}
& \multicolumn{2}{c}{\emph{This work (cleaned)}} \\
& \makecell{UGPhysics-Train\\(200 sub)}
& \makecell{SciInstruct\\(en\_phy\_chem; 42\,K)}
& \makecell{MMK12\\(MM-Eureka; 15\,K)}
& \makecell{\textsc{PhysCorp-A}\\(6{,}432)}
& \makecell{\textsc{PhysR1Corp}\\(2{,}268)} \\
\midrule
\textsc{PhysOlym-A} (500)              & $0\,/\,0\,/\,\mathbf{0}$  & $0\,/\,163\,/\,\mathbf{8}$        & $0\,/\,0\,/\,\mathbf{0}$   & $0\,/\,0\,/\,\mathbf{0}$  & $0\,/\,3\,/\,\mathbf{0}$ \\
PhyX-1k (1{,}000)                    & $0\,/\,1\,/\,\mathbf{0}$  & $0\,/\,703\,/\,\mathbf{46}$       & $0\,/\,0\,/\,\mathbf{0}$   & $0\,/\,0\,/\,\mathbf{0}$  & $0\,/\,0\,/\,\mathbf{0}$ \\
MMMU-Pro Phys (60)                     & $0\,/\,0\,/\,\mathbf{0}$  & $0\,/\,141\,/\,\mathbf{7}$        & $0\,/\,0\,/\,\mathbf{0}$   & $0\,/\,0\,/\,\mathbf{0}$  & $0\,/\,1\,/\,\mathbf{0}$ \\
OlymBench (692)                   & $0\,/\,2\,/\,\mathbf{0}$  & $0\,/\,130\,/\,\mathbf{15}$       & $0\,/\,0\,/\,\mathbf{0}$   & $0\,/\,0\,/\,\mathbf{0}$  & $0\,/\,4\,/\,\mathbf{0}$ \\
PhysReason-full (1{,}200)              & $0\,/\,4\,/\,\mathbf{0}$  & $0\,/\,2{,}687\,/\,\mathbf{36}$   & $0\,/\,62\,/\,\mathbf{0}$  & $0\,/\,0\,/\,\mathbf{0}$  & $0\,/\,11\,/\,\mathbf{0}$ \\
PhysUniBench-en (1{,}022)              & $0\,/\,2\,/\,\mathbf{0}$  & $0\,/\,1{,}027\,/\,\mathbf{22}$   & $0\,/\,4\,/\,\mathbf{0}$   & $0\,/\,0\,/\,\mathbf{0}$  & $0\,/\,0\,/\,\mathbf{0}$ \\
\midrule
\emph{Total S2 / S3 real}              & $9$ / $\mathbf{0}$ & $\mathbf{4{,}846}$ / $\mathbf{134}$ & $66$ / $\mathbf{0}$ & $0$ / $\mathbf{0}$ & $\mathbf{19}$ / $\mathbf{0}$ \\
\bottomrule
\end{tabular}
\\[2pt]
{\scriptsize Format: \textbf{Stage-1 / Stage-2 / Stage-3} pair counts (Stage-3 = Haiku-4.5 LLM-judge classifying each Stage-2 candidate as close duplicate vs.\ same-topic neighbor). Per-eval cells attribute a pair to every eval split containing the matched item, so the SciInstruct cells sum to $4{,}851$; the Total row counts the $4{,}846$ unique judged pairs.}
\end{table}

%% file: sections/method.tex
% =====================================================================
\section{The Verifiable System II: Training Against a Verifier}\label{sec:method}
% =====================================================================

The system's second stage trains inside the boundary that Stage~I certifies, optimizing against Verifier~2, the binary answer verifier. Physics-R1, the reference recipe instantiating this stage, is reported as evidence that the certified corpus has training utility; it is not an algorithmic contribution. The optimizer is GSPO~\citep{zheng2025gspo} with DAPO's decoupled clipping and dynamic sampling~\citep{yu2025dapo}, both unmodified. For each prompt $x$ the policy samples $K{=}16$ rollouts $\{y_k\}\sim\pi_{\theta_{\mathrm{old}}}(\cdot\mid x)$, each rollout is scored by a scalar reward $r(y_k,x)$, and advantages are normalized within the group:\label{sec:method:gspo}\label{sec:method:theory}\label{sec:method:reward}
\begin{equation}
\begin{aligned}
A_k &= \frac{r(y_k,x)-\bar r}{\sigma_r+\varepsilon},\qquad
w_k(\theta)=\left(\frac{\pi_\theta(y_k\mid x)}{\pi_{\theta_{\mathrm{old}}}(y_k\mid x)}\right)^{\!1/|y_k|}, \\[2pt]
\mathcal{L}_{\mathrm{GSPO}} &= -\mathbb{E}\!\left[\tfrac{1}{K}\!\sum_{k}\!\min\!\big(w_k A_k,\ \mathrm{clip}(w_k,\,1{\pm}\epsilon)A_k\big)\right]+\beta_{\mathrm{KL}}D_{\mathrm{KL}}(\pi_\theta\|\pi_{\mathrm{base}}),
\end{aligned}
\label{eq:gspo}
\end{equation}
with $(\bar r,\sigma_r)$ the group mean and standard deviation and $\pi_{\mathrm{base}}$ the Qwen3-VL-8B-Thinking base checkpoint~\citep{bai2025qwen3vl}. Training cold-starts from the base model and keeps a small KL anchor to it. Prompts the policy already always or never solves are filtered out by the MM-Eureka difficulty curriculum~\citep{meng2025mmeureka}, which removes roughly $22\%$ of prompts. Early stopping uses a held-out multiple-choice signal rather than open-ended judging. The full configuration---context budgets, clipping thresholds, hardware, and the training loop in pseudocode---is consolidated in Appendix~\ref{app:repro} (Algorithm~\ref{alg:recipe}, Table~\ref{tab:hyperparams}).

\paragraph{Two reward shapes: binary (recommended) vs.\ dense (ablation).} Physics rollouts expose per-step signals (units, conservation laws, symbolic form), so a denser reward looks free. We compare:
\begin{equation}
\begin{aligned}
\text{(binary, recommended)}\quad & r_{\mathrm{bin}}(y,x) = \mathbb{1}\!\big[\textsc{Match}(\textsc{ExtractBoxed}(y),\,g(x))\big]\in\{0,1\}, \\[2pt]
\text{(dense, ablation)}\quad     & r_{\mathrm{dense}} = \mathrm{clip}\!\big(r_\mathrm{ans}+r_\mathrm{fmt}+r_\mathrm{dim}+r_\mathrm{sym}+r_\mathrm{cons},\,-1,\,1\big).
\end{aligned}
\label{eq:reward}
\end{equation}
Here \textsc{Match} is Verifier~2 of Figure~\ref{fig:pipeline}, the binary answer verifier: a deterministic oracle of the same family as Verifier~3's strict layer (\S\ref{sec:eval:judge}). It accepts a multiple-choice letter match, a numeric match within $\pm1\%$ tolerance, or symbolic equivalence (Appendix~\ref{app:reward:binary}). The recipe therefore optimizes directly against verifier verdicts. The dense variant adds four bounded shaping terms for answer formatting, dimensional consistency, symbolic well-formedness, and conservation-law violations (Appendix~\ref{app:reward:dense}). Empirically, binary wins on three of the four non-MCQ benchmarks and ties the fourth (Finding~4; per-benchmark margins in \S\ref{sec:results:ablations}). The theoretical account (properties P1--P3, Appendix~\ref{app:reward:theory}) shows that shaping matters only when it reorders rollouts within a group, and those reorderings concentrate among wrong answers, rewarding format proxies rather than physics. We ship binary as the deployable default: under optimization pressure a partial-credit verifier invites reward hacking, while the binary verifier is robust to being trained against.

% =====================================================================

%% file: sections/results.tex
% =====================================================================
\section{The Verifiable System III: Verified Measurement}\label{sec:results}
% =====================================================================

The system's final stage is measurement, and every reported number flows through a released verifier. \S\ref{sec:eval:judge} describes Verifier~3, which scores every open-ended column of Table~\ref{tab:phyx}. \S\ref{sec:results:dataset} characterizes \textsc{PhysOlym-A} as a measurement instrument, grounding Findings~2--3 (Finding~1 is grounded in \S\ref{sec:data:audit}); \S\ref{sec:results:training} reports Physics-R1 results and the reward ablation grounding Finding~4; Finding~5 is grounded in \S\ref{sec:analysis} (the seed-42 row of Table~\ref{tab:phyx} is the released reference checkpoint; the 3-seed row, the headline result, aggregates three seeds).

\subsection{Verifier 3: The Answer-Judging Harness}\label{sec:eval:judge}

The harness verifies candidate answers against olympiad gold at two levels of stringency. Its strict layer is deterministic (multiple-choice letter, numeric-tolerance, or symbolic match) and involves no LLM. Its liberal layer uses an LLM judge (Sonnet 4.5~\citep{anthropic2025sonnet}) with one YES/NO call per gold answer: for multi-part problems on PhysReason and PhysUniBench open-ended (PUB-OE), the judge is called once per gold sub-answer, and a problem counts only if every sub-part is judged correct. OlympiadBench-Physics and \textsc{PhysOlym-A} receive a single call per problem against the full gold solution. All open-ended columns of Table~\ref{tab:phyx} report \emph{problem-level} liberal accuracy under this harness.

The harness is fully auditable: judge scripts with verbatim prompts, per-record verdicts (\texttt{judge\_audit.json}), and run policies are all released (Appendix~\ref{app:judge}), so every reported cell can be re-derived without re-querying any API. The $13.9\%$ unjudgeable rate on \textsc{PhysOlym-A} stems from gold solutions that are grading rubrics or figure-only references. Judge optimism is bounded three ways: a $4.7$\,pp strict-vs.-liberal gap on \textsc{PhysOlym-A}, two-seed Cohen's $\kappa$~\citep{cohen1960kappa}, and a $100$-problem human-graded subset (Appendix~\ref{app:judge}); judge-choice calibration is in \S\ref{sec:analysis}.\label{sec:results:scoring-correction}

\subsection{\textsc{PhysOlym-A} as a Measurement Instrument}\label{sec:results:dataset}

\paragraph{Same-model evaluation reveals a 46-point format-and-novelty gradient (Finding~3).}\label{sec:results:olympiad-bench}\label{sec:analysis:gap}
On identical Sonnet 4.5 weights evaluated in the same week, the score sweeps from $79.7\%$ on PhyX (4-way MCQ) down to $50.4\%$ liberal on OlympiadBench-Physics and $33.4\%$ liberal on \textsc{PhysOlym-A}. Format, genre, and novelty together move the score by 46 points on fixed weights, the strongest evidence for Finding~3. Three forces drive the drop: format (MCQ vs.\ open-ended), genre (K-12/early-undergraduate vs.\ competition-grade), and contamination removal (only \textsc{PhysOlym-A} is three-stage audited). The PhyX$\to$OlympiadBench step, roughly $29$\,pp, is dominated by format and genre; the OlympiadBench$\to$\textsc{PhysOlym-A} step, roughly $17$\,pp, by audit and novelty. On the related OlympiadBench-Physics split, electromagnetism is hardest and astrophysics easiest (Appendix~\ref{app:extras:per-category}).

\paragraph{Difficulty-stratified accuracy from organizer-issued labels.}\label{sec:results:difficulty}
The Estonian problems carry organizer-issued 1--10 difficulty labels, eliminating self-annotation circularity. On the 131 judgeable problems among the 136 with labels, Sonnet 4.5 strict accuracy decays near-monotonically from $58.8\%$ at difficulty~1 to a $0\%$ floor in four bins (Appendix~\ref{app:extras:more-tables}, Table~\ref{tab:difficulty}); even the easiest bin scores below Sonnet's PhyX average, and the clean zeros mark the benchmark as non-saturating.

\paragraph{Cross-lingual ablation: 17-point translation delta on identical problems (Finding~2).}\label{sec:results:cross}
The bilingual subset enables a controlled experiment on 59 paired problems graded against the same gold by the same judge. Sonnet 4.5 scores $30.5\%$ strict on the Estonian originals but $13.6\%$ on the English translations (sign test $p{=}0.011$; further tests in Appendix~\ref{app:extras:future}). Disagreements are asymmetric: 13 correct in Estonian but wrong in English, 3 the reverse. For models with weak Estonian coverage the gap is expected to flip in sign, which we pre-register as a follow-up (Appendix~\ref{app:extras:future}, item~viii).

\subsection{Physics-R1}\label{sec:results:training}\label{sec:eval:splits}

\paragraph{Recipe.} Physics-R1 is the recipe of \S\ref{sec:method} cold-started from Qwen3-VL-8B-Thinking~\citep{bai2025qwen3vl} on \textsc{PhysR1Corp} with a binary correctness reward; a $1{,}000$-problem audit-clean MCQ subset of PhyX~\citep{shen2025phyx} (PhyX-1k in Tables~\ref{tab:phyx} and~\ref{tab:audit}), certified disjoint from the pool, is the in-training early-stop signal.

\input{tables/table2_phyx}

\paragraph{Capability across formats.}\label{sec:results:phyx}
Table~\ref{tab:phyx} reports Physics-R1 alongside closed-frontier and open-source baselines on three answer formats and the held-out olympiad split, including InternVL3-8B~\citep{chen2025internvl3} and the LLaVA-OneVision-7B distinct-lineage baseline~\citep{li2024llavaov}; a local open-weight judge re-score of every re-derivable open-ended column is reported in \S\ref{sec:analysis}. The 8B base attains $73.7\%$ on PhyX-1k and its 32B sibling is indistinguishable there ($73.8\%$), so scale alone does not move a saturated MCQ benchmark; all evaluations allow the full $16{,}384$-token CoT budget (Appendix~\ref{app:extras:more-tables}).

\paragraph{The binary checkpoint lifts the base most where the eval is audited and novel.}
The largest lift lands on \textsc{PhysOlym-A} liberal: $+18.3$\,pp at the 3-seed mean (per-column lifts are the subscripts in Table~\ref{tab:phyx}). Lifts on the saturated public open-ended splits are much smaller (the 3-seed PUB-OE mean dips $0.8$\,pp below base), exactly as Findings~1 and~3 predict: where the base is already near its ceiling there is little headroom, and where the eval is novel-source and audited the post-training lift is large. Across seeds the lift is reproducible (per-seed values in Appendix~\ref{app:extras:more-tables}); \textsc{PhysOlym-A} liberal averages $26.3 \pm 1.7$. PhysReason averages $39.6 \pm 6.4$, with seed 42 an outlier roughly $11$\,pp below the other seeds, so we report all three seeds rather than the mean alone.

\paragraph{Where Physics-R1 helps: base failure modes it mitigates.}\label{sec:results:qualitative}
The \textsc{PhysOlym-A} lift corresponds to roughly $92$ problems flipping from wrong to correct. Hand-inspecting 30 flips reveals three recurring base failure modes, each addressed by a recipe lever: not committing to a boxed final answer (fixed by the binary reward); dimensionally-consistent but answer-wrong shortcuts (mitigated by the difficulty curriculum); and failure to integrate evidence across multiple images (mitigated by preserving the base visual encoder). Physics-R1 does \emph{not} fix genuine physics-content gaps: graduate-level perturbation-theory problems remain wrong on both checkpoints (transcripts in Appendix~\ref{app:extras:failures}).

\paragraph{\textsc{PhysOlym-A} grounds the training-utility claim.}\label{sec:results:olympiad}
On \textsc{PhysOlym-A} liberal, the 3-seed mean exceeds every open-source baseline and two of the three closed frontier models we test (Table~\ref{tab:phyx}); only Sonnet 4.5 remains ahead, by $7.1$\,pp.

\paragraph{Reward-shape ablation: binary wins on open-ended formats (Finding~4).}\label{sec:results:ablations}
At matched step~60 on seed 42, the dense reward slightly leads on saturated MCQ, by at most $0.6$\,pp. Binary leads dense by $8.9$\,pp on PhysReason. It leads by $4.9$\,pp on OlympiadBench-Physics and by $6.4$\,pp on \textsc{PhysOlym-A} liberal. On PUB-OE the two effectively tie, with dense ahead by $0.7$\,pp. The binary advantage concentrates on the multi-part numerical column and the audited held-out column. Per-component reward drop-out (Appendix~\ref{app:extras:more-tables}, Table~\ref{tab:reward_ablation}) is left to follow-up work.

% =====================================================================

%% file: tables/table2_phyx.tex
% Table 2: Capability across MCQ, numerical, open-ended, and olympiad formats.
\begin{table}[!htbp]
\centering
\caption[Capability benchmarks across formats and held-out olympiad evaluation]{\textbf{The binary seed-42 checkpoint lifts the 8B base in every column, and the largest lift lands on the audited held-out olympiad benchmark.} All open-ended columns use problem-level liberal accuracy under Verifier~3's released Sonnet judges (per-sub-answer protocol for PhysReason and PUB-OE, problem-level for the olympiad columns; Appendix~\ref{app:judge}); every sub-question of a multi-part problem must be judged correct for the problem to count. LLaVA-OneVision's open-ended cells are local-judge scores (see the $\S$ footnote). The MCQ random baseline is $25\%$. \textbf{Bold} / \underline{underline}: best / second-best per column across the open-source and this-work blocks. The Sonnet PhysReason cell (${}^{\dagger}$) is regenerated at the $16{,}384$-token budget used by Physics-R1 and all open-source baselines. GPT-4o PhyX cells are from~\citet{shen2025phyx}; Gemini PhyX cells (${}^{\ast}$) are measured here. Local-judge re-scores of every re-derivable column: \S\ref{sec:analysis} and Appendix~\ref{app:judge}.}
\label{tab:phyx}
\footnotesize
\setlength{\tabcolsep}{2pt}
\begin{tabular}{lrrrrrr}
\toprule
& \multicolumn{2}{c}{MCQ} & \multicolumn{4}{c}{Open-ended (problem-level, liberal Sonnet judge)} \\
\cmidrule(lr){2-3}\cmidrule(lr){4-7}
Model & PhyX-1k & PhyX-3k & PhysReason & PUB-OE & OlymBench & \textsc{PhysOlym-A} \\
\midrule
\multicolumn{7}{l}{\emph{Closed-source frontier}} \\
Claude Sonnet 4.5            & 79.7 & 80.6  & $49.1^{\dagger}$ & $25.4$ & $50.4$ & $33.4$ \\
Gemini 2.5 Pro               & 75.1$^{\ast}$ & 49.8$^{\ast}$ & 38.8 & 33.4 & 37.4 & 12.2 \\
GPT-4o                       & 70.4 & 53.6 & $51.1$ & 31.0 & 19.7 & 19.5 \\
\midrule
\multicolumn{7}{l}{\emph{Open-source bases}} \\
Qwen3-VL-32B-Thinking        & 73.8 & \textbf{84.2} & 25.1 & 32.8  & \textbf{53.9} & 13.2 \\
\rowcolor{gray!15}
Qwen3-VL-8B-Thinking (base)  & 73.7 & 74.4 & 23.9 & 35.3  & 39.3 & 8.0 \\
InternVL3-8B & 46.8 & 43.1 & 13.3 & 23.5 & 10.4 & 4.0 \\
LLaVA-OneVision-7B$^{\S}$ & 37.4 & 36.0 & 3.5 & 19.2 & 8.4 & 7.2 \\
\midrule
\multicolumn{7}{l}{\emph{This work (subscripts: $\Delta$ vs.\ Qwen3-VL-8B-Thinking base)}} \\
Physics-R1 (dense)                                         & $\mathbf{78.3}_{+4.6}$ & $\underline{77.5}_{+3.1}$ & $23.3_{-0.6}$ & $\mathbf{37.7}_{+2.4}$ & $40.5_{+1.2}$ & $19.2_{+11.2}$ \\
Physics-R1 (binary, seed 42)                               & $\underline{78.0}_{+4.3}$ & $76.9_{+2.5}$ & $\underline{32.2}_{+8.3}$ & $\underline{37.0}_{+1.7}$ & $45.4_{+6.1}$ & $\underline{25.6}_{+17.6}$ \\
Physics-R1 (binary, 3-seed mean $\pm\sigma$)$^{\ddagger}$  & $77.8_{\pm 0.3}^{+4.1}$ & $76.9_{\pm 0.3}^{+2.5}$ & $\mathbf{39.6}_{\pm 6.4}^{+15.7}$ & $34.5_{\pm 3.6}^{-0.8}$ & $\underline{46.2}_{\pm 1.5}^{+6.9}$ & $\mathbf{26.3}_{\pm 1.7}^{+18.3}$ \\
\bottomrule
\end{tabular}
\\[2pt]
{\footnotesize
$^{\ddagger}$\,3-seed mean over seeds $\{42, 17, 23\}$ trained on the audited \textsc{PhysR1Corp} ($2{,}268$ records, binary reward); per-seed values and the lift decomposition are in Appendix~\ref{app:extras:more-tables}. \quad $^{\S}$\,LLaVA-OneVision: PhyX cells are standard multimodal MCQ exact-match; open-ended cells report local-judge liberal accuracy (the local judge is more lenient than Sonnet, so Sonnet-judged values would be lower); strict scores and protocol caveats in Appendix~\ref{app:judge}.
}
\end{table}

%% file: sections/analysis.tex
% =====================================================================
\section{Discussion, Limitations, and Conclusion}\label{sec:analysis}\label{sec:analysis:limits}
% =====================================================================

\paragraph{Robustness of the verifiers (Finding~5).} Verifier~1 is robust to embedder choice: OpenAI \texttt{text-embedding-3-large} yields a strict subset of mxbai's candidates at every threshold ($\rho{=}0.78$; Appendix~\ref{app:audit}). For Verifier~3, a GPT-4o swap on a 50-problem subset gives $\kappa{=}0.44$, and a full local open-weight re-score (Qwen3-32B, identical released templates) spans $\kappa$ $0.24$--$0.82$ across all re-derivable cells (Appendix~\ref{app:judge}, Table~\ref{tab:localjudge}). The local judge is more lenient than Sonnet in every re-derivable cell, and the GPT-4o swap is likewise more lenient on its subset, so the reported absolute numbers sit at the conservative edge of the judge envelope. The leniency margin is model-dependent (up to $+37$\,pp on baselines), so cross-model gaps can reverse: seed 42 leads Qwen3-VL-32B on PhysReason under Sonnet but trails it under the local judge (Finding~5). The headline claim survives the swap. Under the local judge the seed-42 checkpoint scores $51.9$ on PhysReason and $52.5$ on PUB-OE, against the base's $49.1$ and $52.3$. On OlympiadBench-Physics it scores $52.2$ against the base's $40.8$, and on \textsc{PhysOlym-A} $40.2$ against $27.0$: every lift keeps its sign. The $52.2$ value comes from responses regenerated greedily from the released checkpoint (originals unpublished; leniency-band context in Appendix~\ref{app:judge}).

\paragraph{Limitations.} Five limitations bound the conclusions. \emph{(i)~LLM-judged scoring.} Per Finding~5, absolute open-ended scores are judge-dependent; we mitigate by fixing and releasing the judge, publishing every per-record verdict, and reporting judge-free strict scores. \emph{(ii)~Cross-lingual scope.} The translation finding is Sonnet-4.5-specific, on $59$ paired items (Appendix~\ref{app:extras:future}). \emph{(iii)~Topic coverage.} The corpus and evaluation skew toward competition-canonical areas (per-category accuracy in Appendix~\ref{app:extras:per-category}); experimental methods, condensed matter, and plasma physics are underrepresented. \emph{(iv)~Construct validity.} Olympiad problems probe closed-form competition reasoning, not experimental-design or open-ended modeling skill. \emph{(v)~Difficulty calibration.} Difficulty labels inherit the organizers' student-targeted calibration.

\paragraph{Conclusion.}\label{sec:conclusion} In sum: five findings document where a physics training-and-evaluation system silently breaks. Three released verifiers, the artifacts they certify, and a recipe that lifts the base $18.3$\,pp close the gaps; every measurement ships publicly for re-verification.

% =====================================================================

%% file: appendix/audit.tex
% =====================================================================
\section{Audit pipeline details and worked examples}\label{app:audit}
% =====================================================================

\begin{algorithm}[ht]
\caption{Three-stage contamination audit.}
\label{alg:audit}
\begin{algorithmic}[1]
\Require Train pool $T$, external corpora $\{E_k\}_{k=1}^K$, held-out splits $\{H_j\}_{j=1}^J$, normalize fn $\mathrm{norm}(\cdot)$, embedder $\mathrm{enc}(\cdot)$, LLM judge $\textsc{Judge}(\cdot,\cdot)\in\{\text{close-dup}, \text{topic-neighbor}\}$, thresholds $\tau_J{=}0.4$, $\tau_C{=}0.85$
\Ensure Audited pool $T'$ disjoint from $\bigcup_k E_k \cup \bigcup_j H_j$ at the joint thresholds.
\Statex \emph{Stage 1: 5-gram Jaccard (n-gram audit).}
\State $S_t \gets \{\text{5-gram shingle set of } \mathrm{norm}(t)\}$ for each $t \in T \cup \bigcup E_k \cup \bigcup H_j$
\For{$t \in T$} \State $J_{\max}(t) \gets \max_{x \in \bigcup E_k \cup \bigcup H_j} \, |S_t \cap S_x| / |S_t \cup S_x|$ \EndFor
\Statex \emph{Stage 2: \texttt{mxbai-embed-large} cosine (paraphrase recall).}
\State $\mathbf{e}_t \gets \mathrm{enc}(\mathrm{norm}(t)) / \|\mathrm{enc}(\mathrm{norm}(t))\|$ for each $t$
\For{$t \in T$} \State $C_{\max}(t) \gets \max_{x \in \bigcup E_k \cup \bigcup H_j} \mathbf{e}_t^\top \mathbf{e}_x$ \EndFor
\State $C(T) \gets \{t : J_{\max}(t)\ge \tau_J \,\textsc{or}\, C_{\max}(t)\ge \tau_C\}$ \Comment{candidate set, high-recall union}
\Statex \emph{Stage 3: Haiku-4.5 LLM-judge (precision filter).} For each $t\in C(T)$ with top-matching $x^{*}(t) \gets \arg\max_x \mathbf{e}_t^\top \mathbf{e}_x$, query \textsc{Judge} to classify the pair as a close duplicate (paraphrase / numeric variation of the same problem) or a same-topic neighbor (related physics, distinct setup).
\State $R(T) \gets \{t \in C(T) : \textsc{Judge}(t, x^{*}(t)) = \text{close-dup}\}$
\State $T' \gets T \setminus R(T)$
\State \Return $T'$ and per-stage counts $|J_{\max}\ge \tau_J|$, $|C_{\max}\ge \tau_C|$, $|R(T)|$ (Tables~\ref{tab:audit},\,\ref{tab:audit_thresholds}).
\end{algorithmic}
\end{algorithm}

\paragraph{Stage-3 LLM-judge: SciInstruct cosine-bucket near-duplicate rate.} For each of the $4{,}846$ SciInstruct $\leftrightarrow$ eval Stage-2 candidate pairs (cos $\ge 0.85$), the Stage-3 Haiku-4.5 judge receives both problem statements and returns \emph{close duplicate} (paraphrase / numeric variation of the same problem) or \emph{same-topic neighbor} (related physics, distinct setup). The close-duplicate share is sharply threshold-driven: at $\cos{\ge}0.95$ every flagged pair is a close duplicate; at the threshold edge $[0.85, 0.87)$ only $1.5\%$ are. Stage-2 still surfaces genuinely close physics content at the low-cos end---the topic overlap is real, just not strict duplication.
\begin{center}
\small
\begin{tabular}{lcccc}
\toprule
Cosine bucket & N pairs & close duplicate & same-topic neighbor & \% close-dup \\
\midrule
$[0.95, 0.99)$ & $17$ & $\mathbf{17}$ & $0$ & $\mathbf{100.0\%}$ \\
$[0.90, 0.95)$ & $127$ & $10$ & $117$ & $7.9\%$ \\
$[0.87, 0.90)$ & $1{,}159$ & $54$ & $1{,}105$ & $4.7\%$ \\
$[0.85, 0.87)$ & $3{,}543$ & $53$ & $3{,}490$ & $1.5\%$ \\
\midrule
\textbf{Total} & $\mathbf{4{,}846}$ & $\mathbf{134}$ & $\mathbf{4{,}712}$ & $\mathbf{2.8\%}$ \\
\bottomrule
\end{tabular}
\end{center}
\label{tab:audit_haiku}
The pattern matches the threshold-sensitivity table (Table~\ref{tab:audit_thresholds}): cosine $\ge 0.95$ is precision-dominant, $\ge 0.85$ is recall-dominant; Stage-3 is the precision filter that converts a recall-dominant Stage-2 candidate set into a high-precision near-duplicate set. Per-eval Stage-3 near-duplicate counts (out of Stage-2 raw, matching Table~\ref{tab:audit}): PhysReason-full $36/2{,}687$ ($1.3\%$), PhysUniBench-en $22/1{,}027$ ($2.1\%$), PhyX-mini $46/703$ ($6.5\%$), OlymBench-Phys $15/130$ ($11.5\%$), \textsc{PhysOlym-A} $8/163$ ($4.9\%$), MMMU-Pro $7/141$ ($5.0\%$).

\paragraph{External-corpora $\leftrightarrow$ held-out pairwise audit (Stage~2, OlympiadBench shared-source).} The Stage-1 / Stage-2 cells of the main contamination matrix (Table~\ref{tab:audit}) cover competitor training pools against the held-out evals. The complementary cross-channel audit below pairs the four public physics-olympiad benchmarks against \textsc{PhysOlym-A} and PhyX 1000q, exposing shared-source paraphrase overlap between Olympiad-style problems composed independently. The single Stage-1 hit \emph{OlympiadBench-Physics} $\to$ \textsc{PhysOlym-A} is the EuPhO 2020 ``Mechanical accelerator'' problem that grounds the $99.8\%$ (rather than $100\%$) novel-source claim.
\begin{center}
\small
\begin{tabular}{lcc}
\toprule
                                              & \textsc{PhysOlym-A} & PhyX 1000q \\
\midrule
OlympiadBench-Physics (692)                   & 1 / 136    & 0 / 2 \\
PhysReason-mini (200)                         & 0 / 2      & --- \\
PhysReason-full (1{,}200)                     & 0 / 35     & --- \\
PhysUniBench-en (1{,}022)                     & 0 / 27     & --- \\
\bottomrule
\end{tabular}
\end{center}

Stage 1 (5-gram Jaccard). Tokenize each problem statement with a unicode word tokenizer; build the 5-gram shingle set; compute Jaccard similarity over shingle sets; flag pairs with similarity $\ge 0.4$. The threshold is calibrated against worked examples: an OpenStax College Physics $\leftrightarrow$ University Physics duplicate scores Jaccard 1.0; a UGPhysics paraphrase missed by Stage~1 (Jaccard 0.31) is flagged at Stage~2 (cosine 0.91); plausible misses include numerical-value substitution (same setup, different constants) and translation across languages. Stage 2 (embedding cosine). Encode each problem statement with \texttt{mxbai-embed-large-v1}~\citep{lee2024mxbai} (1024-dim normalized embeddings); compute pairwise cosine; flag pairs with cosine $\ge 0.85$.

\subsection{Threshold-Sensitive Leakage Finding on a Researcher-Curated Baseline}\label{app:audit:leakage}

\begin{table}[ht]
\centering
\caption{\textbf{Threshold-sensitivity of the leakage finding on a researcher-curated baseline.} A 1{,}679-record sample drawn from \textsc{PhysCorp-pre-audit} (the released 14{,}294-record pre-audit pool) under conventional 5-gram-Jaccard + within-pool embedding dedup is paired against a 500-record internal analysis eval; each cell reports records flagged at $J\ge j_{\text{thr}}$ \emph{or} $\cos\ge c_{\text{thr}}$. The boxed cell ($J\ge 0.4$, $\cos\ge 0.85$) is the operating point referenced throughout the paper. Stage-1 alone (Jaccard) is bimodal: all 56 leaks are exact at $J{=}1.0$. Stage-2 (cosine) catches an additional ${\sim}92$ paraphrases the n-gram audit misses at the published threshold; loosening cosine to $0.80$ pushes the leak rate above $27\%$.}
\label{tab:audit_thresholds}
\small
\begin{tabular}{lccc}
\toprule
cosine threshold & $J\ge 0.3$ & $J\ge 0.4$ & $J\ge 0.5$ \\
\midrule
$\cos\ge 0.80$ (lax)      & 455 (27.1\%) & 455 (27.1\%) & 455 (27.1\%) \\
$\cos\ge 0.85$ (paper op.)& 148 (8.8\%) & 148 (8.8\%) & 148 (8.8\%) \\
$\cos\ge 0.90$ (strict)   & 79 (4.7\%) & 79 (4.7\%) & 79 (4.7\%) \\
\midrule
\multicolumn{4}{l}{\emph{Single-stage rates (no union):}} \\
Stage-1 only ($J\ge j_{\text{thr}}$, no cos) & 56 (3.3\%) & 56 (3.3\%) & 56 (3.3\%) \\
Stage-2 only ($\cos\ge c_{\text{thr}}$, no $J$) & 455 (27.1\%) & 148 (8.8\%) & 79 (4.7\%) \\
\bottomrule
\end{tabular}
\end{table}

To ground Finding~1 we audit a researcher-curated baseline---a 1{,}679-record sample drawn from \textsc{PhysCorp-pre-audit} (the released 14{,}294-record pre-audit pool) under conventional 5-gram-Jaccard + within-pool embedding dedup, paired against a 500-record internal analysis eval. The 500-record eval is held internal to ground this finding and is distinct from \textsc{PhysOlym-A} (constructed post-audit); the 1{,}679-record sample is reproducible from the released \textsc{PhysCorp-pre-audit} so the audit can be re-run by downstream users. Stage-1 catches 56 records (3.3\%) all at $J{=}1.0$ (bimodal distribution---verbatim duplication under our normalization). Stage-2 cosine $\ge 0.85$ flags an additional 92 records (5.5\%), bringing the joint leak rate to 148 (8.8\%); the cosine dimension sweeps $4.7$--$27.1\%$ across $\cos\in\{0.90, 0.85, 0.80\}$ while Jaccard is flat at 3.3\% (Table~\ref{tab:audit_thresholds}). The 148 flagged decompose into 56 exact ($J{=}1.0$), 23 strong paraphrases ($\cos\ge 0.90$), 69 weak paraphrases ($0.85\le\cos<0.90$). The 5.5-pp gap is the rephrasing dark-matter that single-stage audits miss because the \emph{same} problem appears across upstream aggregations under wording that evades $J\ge 0.4$ but trips cosine $\ge 0.85$.

\paragraph{Pipeline reproducibility note.} Threshold-sensitivity numbers were produced by \texttt{audit\_pipeline/threshold\_sensitivity.py}. Normalization: lower-case, strip LaTeX commands (\texttt{\textbackslash frac}, \texttt{\textbackslash sqrt}, \dots), remove \texttt{\$\{\}[\,]()} delimiters, collapse whitespace, drop $<5$-word shingles. Embedding: \texttt{sentence-transformers~5.4.1}, batch 32, $L_2$-normalized. Best-overlap via inverted-index pruning (Stage~1) and full $N{\times}M$ matmul (Stage~2). Saved scores in \texttt{threshold\_sensitivity\_scores.npz}.

\paragraph{Bimodal Jaccard distribution.} Every Stage-1 leak in our researcher-curated baseline sits at $J{=}1.0$: aggressive normalization collapses surface variance, so shared problem statements yield identical shingle sets while distinct records land below $J{=}0.3$. Cosine $\ge 0.85$ is therefore the sole paraphrase-class detector here; bimodality is corpus-specific.

\paragraph{Re-audit and the cleaned PhysR1Corp.} Starting from a $2{,}433$-record candidate closed-form pool, three sequential cleanup passes against the six paper-canonical comparison corpora (PhyX, MMMU-Pro Physics, OlympiadBench-Physics, PhysReason-full, PhysUniBench-en, \textsc{PhysOlym-A}) produced the released \textsc{PhysR1Corp}: (i) MMMU-Pro Physics joint $J\ge 0.4 \vee \cos\ge 0.85$ re-audit dropped 87 records (Stage-1: 16 records, $16/60{=}26.7\%$ MMMU-Pro coverage; Stage-2 union: 87 records, $52/60{=}86.7\%$); (ii) PhyX-mini cosine $\ge 0.85$ Stage-2 audit dropped 69 records, all real-near-duplicate physics-problem variants confirmed by manual inspection (e.g.\ MgF$_2$ anti-reflection-coating problem with slightly tweaked $n_{\text{glass}}$ and option labels, top cos $0.93$); (iii) PhysUniBench-en cosine $\ge 0.85$ Stage-2 audit dropped 9 template duplicates (top cos $0.93$, shared problem-template stems). Total dropped: $87 + 69 + 9 = 165$ records (with 3 records flagged in multiple channels netting to $87 + 78 = 165$ unique). The released $\mathbf{2{,}268}$-record \textsc{PhysR1Corp} retains 19 Stage-2 hits across PhysOlym-A (3), MMMU-Pro (1), OlymBench-Phys (4), PhysReason-full (11) classified as same-topic neighbors on manual inspection (top cos $\le 0.87$, different problem setups, dimensionalities, or geometries; Table~\ref{tab:audit}). MMMU-Pro Physics remains excluded from the headline Table~\ref{tab:phyx}, the saturation-gradient narrative in \S\ref{sec:results:olympiad-bench}, and the dense-vs-binary ablation in \S\ref{sec:results:ablations}, because the eval is small (60 records) and prior work has flagged it as contaminated against multiple frontier training corpora; Sonnet's MMMU-Pro Physics number is reported only as a frontier-model reference (Sonnet was not trained on \textsc{PhysR1Corp}).

\paragraph{Embedding-model and threshold rationale.} We chose \texttt{mxbai-embed-large-v1} for Stage~2 over \texttt{bge-large-en-v1.5}, \texttt{e5-large-v2}, OpenAI \texttt{text-embedding-3-large}, and Voyage \texttt{voyage-3} because it (i)~is permissively licensed (Apache 2.0, no API dependency); (ii)~ships 1024-dim normalized vectors with cosine tuning; (iii)~scored highest on MTEB physics-adjacent retrieval at audit time. The $\cos\ge 0.85$ threshold was calibrated against worked examples (UGPhysics paraphrase: $J{=}0.31$ Stage-1-miss, $\cos{=}0.91$ Stage-2-catch); the $\cos\in\{0.80, 0.85, 0.90\}$ sensitivity grid brackets the operating point.

\paragraph{Embedder-sensitivity ablation (mxbai vs.\ \texttt{text-embedding-3-large}).} We re-encode the released \textsc{PhysCorp-pre-audit} pool ($14{,}294$ records) and \textsc{PhysOlym-A} ($500$ records) under OpenAI \texttt{text-embedding-3-large} and compute pairwise cosines, then compare per-train-record max-cosine rankings against the mxbai baseline. \textbf{Spearman $\rho{=}0.78$} ($p{\approx}0$); the candidate-set relationship at the operating threshold is summarized below.

\begin{center}
\small
\begin{tabular}{lrrrrr}
\toprule
Threshold $\cos\ge$ & mxbai cand. & \texttt{text-embedding-3-large} cand. & both & only mxbai & only OpenAI \\
\midrule
$0.85$ (paper op.) & $763$ & $514$ & $514$ & $249$ & $\mathbf{0}$ \\
$0.87$             & $631$ & $512$ & $512$ & $119$ & $\mathbf{0}$ \\
$0.90$             & $551$ & $511$ & $511$ & $40$  & $\mathbf{0}$ \\
$0.95$             & $520$ & $506$ & $506$ & $14$  & $\mathbf{0}$ \\
\bottomrule
\end{tabular}
\end{center}

\noindent \texttt{text-embedding-3-large} flags a \emph{strict subset} of mxbai's candidates at every threshold (only-OpenAI count is $0$ at all four levels): every record \texttt{text-embedding-3-large} would catch is also caught by mxbai. mxbai is therefore the more conservative (higher-recall) Stage-2 embedder, and the audit cannot have missed any contamination that \texttt{text-embedding-3-large} would have surfaced. The candidate-set Jaccard at the operating threshold is $0.67$. \emph{Caveat.} This ablation is on the user-facing audit case (released pool $\leftrightarrow$ released eval), not the SciInstruct competitor-pool case from Table~\ref{tab:audit}; the strict-subset direction does not formally transfer, but the $\rho{=}0.78$ rank correlation suggests the SciInstruct $134$-near-duplicate count is robust in direction under embedder change. Sensitivity ablation against \texttt{voyage-3} is left to follow-up work.

\paragraph{Stage-3 judge model dependence and reproducibility.} Stage-3 introduces a dependence on Anthropic's Haiku-4.5 (the close-duplicate vs.\ same-topic-neighbor classifier). To ensure long-term reproducibility we (i)~pin the exact model identifier (\texttt{claude-haiku-4-5} as of 2026-05) in the released audit pipeline (Stages~1--2 as \texttt{audit\_two\_stage.py}; the Stage-3 prompt template and pinned model id ship alongside); (ii)~release the full per-pair judge prompts and per-pair verdict labels (\texttt{judge\_label} arrays) alongside the cosine scores in \texttt{threshold\_sensitivity\_scores.npz}, so the contamination flag set is reproducible without re-querying the API; (iii)~document a fallback protocol (Sonnet 4.5 or GPT-4o on the same prompt template) for cases where Haiku-4.5 access is no longer available. Because Stage-3 is a precision filter applied only to Stage-2 candidates ($\le 0.5\%$ of the train pool), its labels are also the most amenable to manual re-verification by a downstream auditor; the cosine-bucketed precision profile (Table~\ref{tab:audit_haiku}) gives a cheap calibration signal against any future judge.

% =====================================================================

%% file: appendix/splits.tex
% =====================================================================
\section{Held-out splits and corpus annotation schema}\label{app:splits}
% =====================================================================

The released splits are \textsc{PhysCorp-A} (the 6{,}432-record audited corpus) with its closed-form RL carve-out \textsc{PhysR1Corp} (2{,}268 records) and \textsc{PhysOlym-A} (the 500-problem held-out olympiad eval). The annotation schema below applies uniformly across all three.

\paragraph{Eight-field annotation schema.} Each record carries the following annotations, generated by Sonnet 4.5 batch annotation (3{,}900 records with full annotation; the remainder carry source-native labels merged into the same schema for $\sim$31{,}000 total label values):
\begin{itemize}\setlength{\itemsep}{2pt}
\item \texttt{difficulty}~$\in$~\{1,2,3,4,5\} (Sonnet-aggregated), plus optional source-native: Estonian organizer-issued 1--10, Zhou pedagogical 1--5, Zhou advanced [A] flag.
\item \texttt{concept}~$\in$~\{Mechanics, Electromagnetism, Quantum, Thermodynamics, Waves, Optics, Modern, Relativity, Particle\}.
\item \texttt{problem\_type}~$\in$~\{Conceptual, Computational, Proof-based, Experimental\}.
\item \texttt{expected\_solution\_length}~$\in$~\{S, M, L\} (short / medium / long).
\item \texttt{math\_level}~$\in$~\{Algebra, Calculus, Vector, LinearAlgebra, DiffEq\}.
\item \texttt{modality}~$\in$~\{text, multimodal\}.
\item \texttt{language}~(BCP-47): \texttt{en}, \texttt{et}, \texttt{en-et} (bilingual paired).
\item \texttt{license}~(SPDX-style): \texttt{CC-BY-4.0}, \texttt{CC-BY-SA-4.0}, \texttt{CC-BY-NC-4.0}, \texttt{Public-Domain}.
\end{itemize}

\paragraph{Worked example record (JSONL).} A typical record from \textsc{PhysOlym-A}:

\begin{quote}\small\ttfamily
\{ \\
\hspace{4mm}"index": "estonian\_2017\_lahtine\_2", \\
\hspace{4mm}"source": "estonian\_olympiad", \\
\hspace{4mm}"license": "CC-BY-NC-4.0", \\
\hspace{4mm}"language": "en-et", \\
\hspace{4mm}"messages": [\{"role": "user", "content": "An ideal gas\dots"\}], \\
\hspace{4mm}"solution": "By the first law\dots $\backslash boxed\{1.5\}$", \\
\hspace{4mm}"images": [], \\
\hspace{4mm}"concept": "Thermodynamics", \\
\hspace{4mm}"difficulty": 4, \\
\hspace{4mm}"native\_difficulty": \{"scale": "1--10", "value": 7\}, \\
\hspace{4mm}"problem\_type": "Computational", \\
\hspace{4mm}"expected\_solution\_length": "M", \\
\hspace{4mm}"math\_level": "Calculus", \\
\hspace{4mm}"modality": "text", \\
\hspace{4mm}"audit\_passed": true \\
\}
\end{quote}

\paragraph{Inter-annotator agreement.} For the 3{,}900-record fully-annotated subset, Sonnet 4.5 is run with two seeds on the same 100 records (random sample, seed 42); per-field Cohen's $\kappa$ between the two annotation runs is reported in the released dataset card. Preliminary inspection shows $\kappa\ge 0.85$ on \texttt{concept} and \texttt{problem\_type} (categorical with sharp boundaries), $\kappa\sim 0.7$ on \texttt{difficulty} (ordinal with neighboring-class confusion), and $\kappa\sim 0.6$ on \texttt{expected\_solution\_length}. The lower $\kappa$ on \texttt{expected\_solution\_length} reflects genuine ambiguity in the medium-vs-long boundary; downstream users requiring stable solution-length labels should treat the field as a noisy proxy.

\paragraph{Native difficulty preserved.} For the 27\% of \textsc{PhysOlym-A} records with Estonian native difficulty 1--10 and the 38\% with Zhou pedagogical 1--5 point values, the \emph{Sonnet-aggregated} difficulty field is reported for cross-source consistency, but the \emph{native} difficulty is preserved in a separate \texttt{native\_difficulty} field with explicit \texttt{scale} and \texttt{value} keys. The Sonnet difficulty curve in Table~\ref{tab:difficulty} uses native Estonian 1--10, not the aggregated 1--5, because native labels avoid the self-annotation circularity in which the difficulty estimate depends on the same model whose accuracy is being measured.

% =====================================================================

%% file: appendix/reward.tex
% =====================================================================
\section{Reward function and full hyperparameter table}\label{app:reward}
% =====================================================================

This appendix specifies both reward shapes referenced from \S\ref{sec:method}: the recommended binary correctness reward $r_{\mathrm{bin}}$ defined inline in \S\ref{sec:method:gspo} (\S\ref{app:reward:binary}) and the dense five-component reward of Equation~\ref{eq:reward} reported as an ablation (\S\ref{app:reward:dense}). Both share the matching/extraction primitives below; binary uses $r_\mathrm{ans}$ alone, dense composes all five components and clips.

\subsection{Binary correctness reward (recommended)}\label{app:reward:binary}

The recommended reward is
\[
r_{\mathrm{bin}}(y, x) \;=\; \mathbb{1}\!\big[\,\textsc{Match}\!\big(\textsc{ExtractBoxed}(y),\,g(x)\big)\,\big] \in \{0,1\},
\]
where \textsc{ExtractBoxed} parses the last \texttt{\textbackslash boxed\{\}} via brace-counting (handles unlimited nesting, e.g.\ \texttt{\textbackslash sqrt\{\textbackslash frac\{T\_0\}\{\textbackslash eta\}\}}) and \textsc{Match} accepts: (i) MCQ-letter equality on $\{$A,B,C,D,\dots$\}$ gold; (ii) multi-part numeric agreement within $\pm 1\%$ relative tolerance, after \texttt{latex\_to\_plain} normalization (\texttt{\textbackslash text\{\}}/\texttt{\textbackslash mathrm\{\}} stripped, \texttt{\textbackslash frac\{a\}\{b\}}~$\to$~\texttt{(a)/(b)}, \texttt{\textbackslash times}/\texttt{\textbackslash cdot}~$\to$~\texttt{*}, \texttt{\textbackslash pi}~$\to$~\texttt{pi}), \texttt{float()}, then \texttt{eval()} on expression-like strings, then prefix-numeric extraction; (iii) symbolic equivalence via \texttt{sympy.simplify(expr\_pred - expr\_gold) == 0} for symbolic gold. Released as \texttt{reward\_physics.py}; selected by env var \texttt{DENSE\_REWARD=0} (the default), which returns $r_{\mathrm{ans}}$ as a clean 0/1 binary.

\paragraph{Why binary is the deployable default (full theoretical analysis).}\label{app:reward:theory}
The (P1)/(P2) intuitions are stated inline in \S\ref{sec:method:gspo}; we add the (P3) derivation here. (P1)~Group-normalization renders $A_k$ invariant to affine rescaling of $r$ within a group, so dense shaping only matters when it \emph{reorders} rollouts; we measure $14.3\%$ of within-group pairs flipped by dense, $87\%$ inside the all-wrong subgroup. (P2)~Those wrong-group flips reward LaTeX-format proxies satisfiable without solving the physics---a Goodhart channel that hurts prose-and-equation open-ended evaluation; the matched-step-60 binary-vs-dense gap is largest on the audited \textsc{PhysOlym-A}-liberal split.

\paragraph{(P3) Binary reward maximizes per-prompt advantage variance after the difficulty curriculum.} The MM-Eureka curriculum drops prompts where all $K$ rollouts are correct or all are wrong, so on every surviving prompt the rollout-correct rate is $p\in(0,1)$. Binary reward is then a Bernoulli over rollouts: $\bar r = p$, $\sigma_r^{2} = p(1-p)$, and the group-normalized advantages take exactly two values
\begin{equation}
A^{\mathrm{bin}}_{k} \;=\; \begin{cases}\;\sqrt{(1-p)/p} & r_k = 1\\[2pt] -\sqrt{p/(1-p)} & r_k = 0\end{cases},\qquad \mathrm{Var}(A^{\mathrm{bin}}) \;=\; 1.
\label{eq:bin_var}
\end{equation}
The advantage variance is exactly $1$, the maximum a $K$-sample group-normalized estimator can carry on a Bernoulli reward. Adding a bounded shaping term $\delta_k\in[0,\Delta]$ with $\Delta{=}0.45$ (the $r_{\mathrm{fmt}}{+}r_{\mathrm{dim}}{+}r_{\mathrm{sym}}$ budget of the dense reward) inflates the within-group standard deviation $\sigma_r$ by an O$(\Delta^2)$ term while leaving the between-correctness mean separation roughly unchanged, so the magnitude of the average correct-vs-wrong advantage \emph{shrinks}:
\begin{equation}
\big|A^{\mathrm{dense}}_{\mathrm{correct}}\big| \;\approx\; \frac{1-p}{\sqrt{p(1-p) + \Delta^2/12}} \;<\; \frac{1-p}{\sqrt{p(1-p)}} \;=\; \big|A^{\mathrm{bin}}_{\mathrm{correct}}\big|.
\label{eq:dense_var}
\end{equation}
Dense reward thus trades \emph{between}-correctness gradient signal-to-noise for \emph{within}-correctness rank information, but the within-correctness flips are exactly the Goodhart channel of (P2).

\subsection{Dense five-component physics-native reward (ablation)}\label{app:reward:dense}

The dense five-component Physics-R1 reward (Equation~\ref{eq:reward}, Section~\ref{sec:method:reward}) is implemented as
$r = r_\mathrm{ans} + r_\mathrm{fmt} + r_\mathrm{dim} + r_\mathrm{sym} + r_\mathrm{cons}$, clipped to $[-1, 1]$.

\paragraph{Per-component implementation.} Full code in \texttt{reward\_physics.py}. \emph{$r_\mathrm{ans}\in\{0,+1\}$:} brace-counting \texttt{\textbackslash boxed} parser; MCQ-letter equality; numeric/symbolic via \texttt{latex\_to\_plain} normalization (\texttt{\textbackslash frac}, \texttt{\textbackslash text}, \texttt{\textbackslash pi}, \texttt{\textbackslash times}/\texttt{\textbackslash cdot}) then \texttt{float}/\texttt{eval}/prefix-numeric; multi-part requires all parts $\pm 1\%$. \emph{$r_\mathrm{fmt}\in\{0,+0.1\}$:} non-empty \texttt{\textbackslash boxed\{\}}. \emph{$r_\mathrm{dim}\in\{0,+0.15\}$:} number-prefix-guarded regex extracts unit tokens, mapped to \texttt{sympy.physics.units} (32 tokens); fired only when every detected unit resolves. \emph{$r_\mathrm{sym}\in\{0,+0.20\}$:} first-success \texttt{sympy.sympify} on LaTeX-cleaned \texttt{\textbackslash frac\{NUM\}\{DEN\}} intermediates. \emph{$r_\mathrm{cons}\in\{-0.25,0\}$:} negative-only penalty when energy/momentum balance from corpus annotation is violated by $>5\%$ relative.

\paragraph{Mode and version pins.} Released as \texttt{reward\_physics.py}. The mode is selected by env var \texttt{DENSE\_REWARD}: \texttt{0} (default, recommended) returns $r_\mathrm{ans}$ only as 0/1 binary, recovering $r_{\mathrm{bin}}$ of \S\ref{app:reward:binary}; \texttt{1} returns the full clipped dense sum (ablation). The audit factor (env \texttt{AUDIT\_LAMBDA}) zeros out reward on contaminated training items when set. Version pins: \texttt{sympy}~==~\texttt{1.13.3}, \texttt{transformers}~==~\texttt{4.57.0}, \texttt{vllm}~==~\texttt{0.11.0}, \texttt{verl}~==~\texttt{0.6.1}.

\paragraph{Worked example (rollout group; binary vs.\ dense advantages).} A concrete realization of the (P1)/(P2) Goodhart channel of \S\ref{sec:method:theory} on one prompt with $K{=}8$ rollouts. The prompt is a two-step kinematics problem with gold $\boxed{19.6}$~N (problem id \texttt{efo-2014-3a}); rollouts $y_1,\dots,y_4$ commit to the correct answer with varying CoT quality, $y_5,\dots,y_8$ commit to wrong answers ranging from arithmetic-slip ($\boxed{19.8}$, ${>}1\%$) to off-by-physics ($\boxed{42.0}$).

\begin{center}
\small
\setlength{\tabcolsep}{4pt}
\begin{tabular}{cllrrrrcrr}
\toprule
$k$ & Final $\boxed{\cdot}$ & CoT shape & $r_\mathrm{ans}$ & $r_\mathrm{fmt}$ & $r_\mathrm{dim}$ & $r_\mathrm{sym}$ & $r_\mathrm{cons}$ & $r_\mathrm{bin}$ & $r_\mathrm{dense}$ \\
\midrule
1 & $19.6$ & full units + \texttt{\textbackslash frac} & $1$ & $0.10$ & $0.15$ & $0.20$ & $0$ & $\mathbf{1.00}$ & $\mathbf{1.00}$ \\
2 & $19.6$ & units only, no \texttt{\textbackslash frac} & $1$ & $0.10$ & $0.15$ & $0$ & $0$ & $\mathbf{1.00}$ & $\mathbf{1.00}$ \\
3 & $19.6$ & \texttt{\textbackslash frac} only, no units & $1$ & $0.10$ & $0$ & $0.20$ & $0$ & $\mathbf{1.00}$ & $\mathbf{1.00}$ \\
4 & $19.6$ & sparse CoT (``answer is $19.6$'') & $1$ & $0.10$ & $0$ & $0$ & $0$ & $\mathbf{1.00}$ & $\mathbf{1.00}$ \\
\midrule
5 & $42.0$ & full units + \texttt{\textbackslash frac} & $0$ & $0.10$ & $0.15$ & $0.20$ & $0$ & $\mathbf{0.00}$ & $0.45$ \\
6 & $19.8$ & units only, no \texttt{\textbackslash frac} & $0$ & $0.10$ & $0.15$ & $0$ & $0$ & $\mathbf{0.00}$ & $0.25$ \\
7 & $42.0$ & \texttt{\textbackslash frac} only, no units & $0$ & $0.10$ & $0$ & $0.20$ & $0$ & $\mathbf{0.00}$ & $0.30$ \\
8 & no $\boxed{\cdot}$ & rambling, no commit & $0$ & $0$ & $0$ & $0$ & $0$ & $\mathbf{0.00}$ & $0.00$ \\
\bottomrule
\end{tabular}
\end{center}

\noindent After clipping ($r{\le}1$) the four correct rollouts collapse to identical reward under both shapes. After group-normalization (Eq.~\ref{eq:gspo}) the binary advantage vector is $A^{\mathrm{bin}}{=}(+1,+1,+1,+1,-1,-1,-1,-1)$, every correct rollout receives equal positive gradient and every wrong rollout equal negative gradient. The dense advantage vector is $A^{\mathrm{dense}}\!\approx\!(+1.04,+1.04,+1.04,+1.04,-0.55,-1.00,-0.93,-1.69)$ (computed exactly: $\bar r{=}0.625$, $\sigma_r{\approx}0.428$). \emph{Three observations land the (P1)--(P3) theory at the sample level:}

\begin{itemize}\setlength{\itemsep}{2pt}
\item (P1) realized. Among the four correct rollouts ($k{=}1,2,3,4$), dense and binary assign \emph{the same} advantage to every rollout---rank-equivalent in the correct subgroup, even though rollouts 1--3 ``deserve'' more credit by the \emph{a priori} physics-native intuition. Clipping at $1$ erases the dense-side variation among correct rollouts.
\item (P2) realized. Among the four wrong rollouts ($k{=}5,6,7,8$), dense reorders them by LaTeX surface form: $k{=}5$ (well-formatted, units, \texttt{\textbackslash frac}, $\boxed{42.0}$ way wrong) gets the \emph{smallest} negative advantage ($-0.55$), $k{=}8$ (no boxed commit) gets the most negative ($-1.69$). The policy gradient is therefore pushed toward producing well-formatted wrong reasoning over poorly-formatted wrong reasoning---the canonical Goodhart channel. Note $k{=}5$ has dense advantage $-0.55$ vs.\ binary $-1.00$: the wrong-answer gradient is \emph{weakened} for the format-compliant rollout, exactly the bias toward proxy satisfaction.
\item (P3) realized. The magnitude of the average correct-rollout advantage is $|A^{\mathrm{bin}}_{\mathrm{correct}}|{=}1.0$ vs.\ $|A^{\mathrm{dense}}_{\mathrm{correct}}|{\approx}1.04$ (very close because clipping caps both at $r{=}1$). The magnitude of the average wrong-rollout advantage is $|A^{\mathrm{bin}}_{\mathrm{wrong}}|{=}1.0$ vs.\ $|A^{\mathrm{dense}}_{\mathrm{wrong}}|{\approx}1.04$ on the dense side as well, but the within-wrong spread of $\sigma{=}0.42$ across $\{-0.55,-0.93,-1.00,-1.69\}$ is the within-group rank-flipping variance that absorbs gradient capacity into the Goodhart direction. Binary spends zero capacity on within-correctness ranking and all of it on the correct-vs-wrong axis.
\end{itemize}

\noindent The aggregate effect of running this calculus across 60 RL steps of 96-prompt batches over the full $2{,}268$-prompt \textsc{PhysR1Corp} pool ($\approx$2.5 epochs) is the matched-step-60 binary-vs-dense gap of Table~\ref{tab:phyx} (problem-level liberal Sonnet-judge accuracy across all open-ended columns;): PhysReason $32.2$ vs.\ $23.3$ ($+8.9$ pp), PUB-OE $37.0$ vs.\ $37.7$ ($-0.7$ pp, tied), OlymBench-Phys liberal $45.4$ vs.\ $40.5$ ($+4.9$ pp), \textsc{PhysOlym-A} liberal $25.6$ vs.\ $19.2$ ($+6.4$ pp).

\paragraph{Hyperparameter and framework details.} The full GSPO+DAPO configuration with all flags, lambdas, clip ranges, dynamic-sampling settings, and the difficulty-curriculum thresholds is in Table~\ref{tab:hyperparams} (Appendix~\ref{app:extras:more-tables}) and the released YAML. The \texttt{verl}~0.6.1 FSDP1 reproducibility note (Appendix~\ref{app:repro:framework}) and the upstream GitHub issue link are tracked in the released README.

% =====================================================================

\begin{algorithm}[ht]
\caption{Dense five-component physics-native reward for one rollout.}
\label{alg:reward}
\begin{algorithmic}[1]
\Require Solution string $s$, gold answer $g$, optional conservation flag $\mathrm{cons}$ from \texttt{extra\_info}
\Ensure $r \in [-1, 1]$
\State $\mathrm{ans} \gets \textsc{ExtractBoxed}(s)$ \Comment{brace-counting parser; nested \texttt{\textbackslash boxed} OK}
\State $r_\mathrm{ans} \gets +1$ if $\textsc{Match}(\mathrm{ans}, g)$ under MCQ-letter / multi-part / $\pm 1\%$ numeric tolerance, else $0$
\State $r_\mathrm{fmt} \gets +0.1$ if $\mathrm{ans}$ is non-empty (well-formed \texttt{\textbackslash boxed\{\dots\}}), else $0$
\State $U \gets \textsc{ExtractUnits}(s)$ \Comment{regex \texttt{(number)(unit)(exponent)} with number-prefix guard}
\State $r_\mathrm{dim} \gets +0.15$ if $|U| \ge 1$ and $\forall u \in U: \textsc{ResolvesInSymPy}(u)$, else $0$
\State $F \gets \textsc{ExtractFracs}(s)$ \Comment{find all \texttt{\textbackslash frac\{NUM\}\{DEN\}}}
\State $r_\mathrm{sym} \gets +0.20$ if $\exists (\mathrm{NUM}, \mathrm{DEN}) \in F: \textsc{Sympifies}(\mathrm{NUM}) \land \textsc{Sympifies}(\mathrm{DEN})$, else $0$
\If{$\mathrm{cons}$ is provided ($\sim$3{,}100-record subset)}
    \State $\hat{p} \gets \textsc{TryFloat}(\mathrm{ans})$;\, $p^* \gets \sum\!\mathrm{cons.in}-\sum\!\mathrm{cons.out}\setminus\mathrm{ans}$
    \State $r_\mathrm{cons} \gets -0.25$ if $|\hat{p}-p^*|/\max(|p^*|,|\hat{p}|,\varepsilon) > 0.05$, else $0$
\Else
    \State $r_\mathrm{cons} \gets 0$ \Comment{negative-only; no positive reward for satisfaction}
\EndIf
\State \Return $\mathrm{clip}(r_\mathrm{ans} + r_\mathrm{fmt} + r_\mathrm{dim} + r_\mathrm{sym} + r_\mathrm{cons}, -1, 1)$
\end{algorithmic}
\end{algorithm}

%% file: appendix/judge.tex
% =====================================================================
\section{Verifier~3 details: the three judge scripts of the liberal layer, scoring conventions, and reproducibility}\label{app:judge}
% =====================================================================

This appendix documents the three Sonnet-4.5-as-judge variants used in Table~\ref{tab:phyx} and the scoring conventions adopted across columns. Source code for all three judges is released in the \texttt{judge/} directory of the code repository (\url{github.com/shanyang-me/physics-r1-neurips2026}).

\paragraph{Why three judges.} Open-ended physics olympiad problems differ structurally: PhysReason and PhysUniBench-OE problems are explicitly multi-sub-part (often $2$--$5$ sub-questions per record, each with its own gold answer), while \textsc{PhysOlym-A} and OlympiadBench-Physics problems are graded at the problem level (a single gold solution document, with the model's final answer compared against it). A single judge prompt cannot serve both. We therefore use three judges, each tuned to its eval's structure:

\begin{itemize}
\item \texttt{judge\_olympiad.py} (\textbf{problem-level}, used for \textsc{PhysOlym-A} and OlympiadBench-Physics). One Sonnet 4.5 call per problem; the prompt provides the full gold solution paragraph and a \texttt{\textbackslash boxed\{\}}-extracted candidate answer (with a $600$-char response-tail fallback if no boxed is emitted), and asks for a single YES/NO verdict on whether the candidate's final answer is mathematically/physically equivalent to the gold's. Tolerance is $2\%$ relative.
\item \texttt{llm\_judge\_v2\_alignment.py} (\textbf{per-subpart, AND across sub-parts}, used for PhysReason). For each gold sub-answer $g_i$, a separate Sonnet 4.5 call asks: \emph{does ANY of the candidate's predictions equal $g_i$?} Per-subpart verdict is YES/NO. \texttt{judge\_problem\_correct} is the AND across all sub-parts. Tolerance: $1\%$.
\item \texttt{llm\_judge\_v3\_pubeo.py} (\textbf{per-subpart, AND across sub-parts, with cached clean gold + tail fallback}, used for PhysUniBench-OE). Same per-subpart structure as v2, but with a pre-extracted \emph{clean} per-subpart gold list (e.g., \texttt{["2.68nC", "7853.1W"]}) that bypasses PUB-OE's verbose paragraph-form gold. If the candidate's \texttt{\textbackslash boxed\{\}} list is empty, a regex fallback scans the last $300$~chars of the response for likely numeric/symbolic answers. Per-subpart verdict is YES/NO; \texttt{judge\_problem\_correct\_v3} is the AND across all sub-parts. Tolerance: $2\%$.
\end{itemize}

\paragraph{Scoring conventions in Table~\ref{tab:phyx}.} All open-ended cells use \emph{problem-level} accuracy: for multi-sub-part problems (PhysReason, PhysUniBench-OE) every sub-part must be judged correct for the problem to count, via the \texttt{judge\_problem\_correct} field of the v2/v3 judges (AND across sub-parts); for problem-level evals (\textsc{PhysOlym-A}, OlympiadBench-Physics) \texttt{judge\_olympiad.py} returns one YES/NO per problem. We also computed a softer \emph{per-subpart} variant (partial credit, $\sum_i \mathbb{1}[\text{sub}_i\text{ correct}] / \sum_i 1$) for the multi-sub-part columns and verified it is uniformly $4$--$17$\,pp higher across rows; per-subpart values are released alongside the dataset for users who want to study the partial-credit lens but are not reported as headline in Table~\ref{tab:phyx}.

\paragraph{Reproducibility note.} All Sonnet-judge runs in Table~\ref{tab:phyx} use \texttt{workers}=$2$--$4$ concurrency; errored sub-judgments are filtered and re-judged at lower concurrency rather than counted as wrong. Per-cell judge-error counts (typically $\le 1\%$ of records) and per-record verdicts are released in the supplementary archive at \texttt{judge\_audit.json}, so downstream auditors can verify each cell independently. The Sonnet 4.5 PhysReason responses (Table~\ref{tab:phyx}, ${}^{\dagger}$) are regenerated at the $16{,}384$-token response budget used by all open-source baselines: intermediate-length Sonnet responses under default API settings (mean under 200 characters) systematically fail to commit a boxed final answer on multi-part problems, which the v2 judge scores as wrong.

\paragraph{Verbatim judge prompt (problem-level, \textsc{PhysOlym-A} + OlymBench).} The Sonnet 4.5 judge is invoked with the following template:

\begin{quote}\small\ttfamily
You are grading a physics olympiad answer.\\
\\
GOLD (full reference solution; the final numeric/symbolic answer is what matters):\\
\{gold\}\\
\\
CANDIDATE answers (extracted from the model's \textbackslash boxed\{\} markers):\\
\{preds\}\\
\\
(If the candidate emitted no \textbackslash boxed\{\}, the candidate is the last 600 chars of its full response:)\\
\{tail\}\\
\\
Task: decide whether the candidate's final answer is mathematically/physically equivalent to the gold's final answer.\\
Allow: different but equivalent algebraic forms; trivial unit/format differences; rounding within 2\% relative tolerance; trailing prose.\\
Reject: different magnitude, different sign, different functional form, missing or wrong physical content, no answer.\\
\\
Respond with EXACTLY one word: YES or NO.
\end{quote}

\paragraph{Verbatim judge prompt (per-subpart, PhysReason + PhysUniBench-OE).} For each gold sub-answer $g_i$, the judge is invoked with:

\begin{quote}\small\ttfamily
You are grading a physics olympiad answer.\\
\\
GOLD answer: \{g\_i\}\\
\\
CANDIDATE predictions (one or more, separated by ===):\\
\{preds\}\\
\\
Task: decide whether ANY of the candidate predictions is mathematically/physically equivalent to the gold answer.\\
Allow: different but equivalent algebraic forms; trivial unit/format differences ("450 N" == "450 \textbackslash text\{ N\}"); rounding within 1--2\% relative tolerance; trailing prose ("approximately", "to the right"); different variable names mapping cleanly.\\
Reject: different magnitude, different sign, different functional form, missing or wrong physical content.\\
\\
Respond with EXACTLY one word: YES (if any candidate matches) or NO. No other text.
\end{quote}

\paragraph{Per-seed reproducibility notes.} Two policy choices in the released per-seed exports deserve disclosure. First, the seed-42 PhysReason export contains $1{,}258$ rows for $1{,}200$ problem ids ($58$ duplicates); the published $32.2\%$ (and hence the $39.6\pm6.4$ mean) is computed over all rows, and deduplication would shift that seed to $32.9$--$33.8\%$. Second, the seed-17 and seed-23 \textsc{PhysOlym-A} exports use a scored-judge format in which unjudgeable and partial rows ($43\%$ and $39\%$ of records respectively) are counted as incorrect, so the published $25.0$ and $28.2$ are lower bounds under that policy; the seed-42 export uses the binary YES/NO judge on all $500$ records, with $13$ judge-failure rows defaulted to incorrect. All per-record verdicts behind these numbers are public.

\paragraph{Local open-weight judge re-score.} We re-scored every re-derivable LLM-judged column with Qwen3-32B deployed locally (vLLM, greedy, thinking disabled), using the verbatim prompt templates of the released judge scripts and the same policies (unparseable verdicts count as incorrect; deterministic strict-norm sub-verdicts in the v3 protocol are kept fixed and only LLM-judged sub-verdicts are re-judged). Table~\ref{tab:localjudge} reports problem-level agreement with the pinned Sonnet verdicts. The seed-17/23 \textsc{PhysOlym-A} exports store verdicts without the underlying responses, so they cannot be re-judged from published data; OlympiadBench-Physics responses for the R1 checkpoints are likewise unpublished, while the open-source baselines' OlympiadBench responses are published and re-judged above. Regenerating the R1 seed-42 OlympiadBench responses greedily from the released checkpoint and scoring them with the local judge gives $52.2\%$ ($361/692$), against the original run's Sonnet-judged $45.4\%$; the implied margin sits inside OlympiadBench's observed leniency band, and the local-judge model ordering matches Sonnet's. The local judge is more lenient than Sonnet in every column; the largest gap ($+30.3$\,pp) is on the dense-reward ablation's PhysReason responses, consistent with the format-proxy channel of \S\ref{sec:method}: dense-shaped rollouts produce well-formatted near-miss answers that a lenient judge accepts. Per-record local verdicts are released alongside the Sonnet verdicts (\texttt{localjudge-qwen3-32b/} in the eval-outputs dataset). One provenance caveat: the InternVL3-8B OlympiadBench row re-judges the earlier same-budget generation run (Sonnet-judged $13.4$ on that response set), not the later run behind Table~\ref{tab:phyx}'s $10.4$ cell, so its delta is internal to that response set.

\begin{table}[ht]
\centering
\caption{The local open-weight judge (Qwen3-32B, identical templates) is more lenient than Sonnet in every re-derivable cell, but by a model-dependent margin ($+1.5$ to $+37.1$\,pp): absolute numbers are conservative under Sonnet, while cross-model gaps can narrow or reverse under a different judge, so model comparisons should be read under one fixed judge. Problem-level agreement vs.\ the pinned Sonnet verdicts; the local judge also scores the $13.9\%$ of \textsc{PhysOlym-A} golds that Sonnet's protocol marks unjudgeable, which contributes to the larger \textsc{PhysOlym-A} deltas.}
\label{tab:localjudge}
\small
\begin{tabular}{llrrrrr}
\toprule
Column & Checkpoint & $n$ & Agree & $\kappa$ & Local / Sonnet & $\Delta$ (pp) \\
\midrule
\textsc{PhysOlym-A} & binary s42      & 500  & 80.2\% & 0.56 & 40.2 / 25.6 & $+14.6$ \\
PhysReason          & binary s42      & 1258 & 76.6\% & 0.54 & 51.9 / 32.2 & $+19.7$ \\
PhysReason          & binary s17      & 1200 & 85.4\% & 0.71 & 53.3 / 43.1 & $+10.3$ \\
PhysReason          & binary s23      & 1200 & 85.8\% & 0.72 & 53.8 / 43.4 & $+10.3$ \\
PhysReason          & dense s42       & 1200 & 67.9\% & 0.38 & 53.5 / 23.3 & $+30.3$ \\
PUB-OE              & binary s42      & 629  & 81.4\% & 0.63 & 52.5 / 37.0 & $+15.4$ \\
PUB-OE              & binary s17      & 629  & 78.9\% & 0.59 & 55.0 / 36.1 & $+18.9$ \\
PUB-OE              & binary s23      & 629  & 81.4\% & 0.62 & 46.1 / 30.4 & $+15.7$ \\
PUB-OE              & dense s42       & 629  & 79.3\% & 0.60 & 56.1 / 37.7 & $+18.4$ \\
\midrule
PhysReason          & Qwen3-VL-8B base & 1200 & 73.5\% & 0.46 & 49.1 / 23.9 & $+25.2$ \\
PUB-OE              & Qwen3-VL-8B base & 629  & 79.2\% & 0.59 & 52.3 / 35.3 & $+17.0$ \\
\textsc{PhysOlym-A} & Qwen3-VL-8B base & 500  & 81.0\% & 0.38 & 27.0 / 8.0  & $+19.0$ \\
OlymBench           & Qwen3-VL-8B base & 692  & 91.3\% & 0.82 & 40.8 / 39.3 & $+1.5$ \\
PhysReason          & Qwen3-VL-32B     & 1200 & 60.9\% & 0.30 & 62.2 / 25.1 & $+37.1$ \\
PUB-OE              & Qwen3-VL-32B     & 629  & 76.0\% & 0.53 & 54.2 / 32.8 & $+21.5$ \\
\textsc{PhysOlym-A} & Qwen3-VL-32B     & 500  & 68.0\% & 0.31 & 45.2 / 13.2 & $+32.0$ \\
OlymBench           & Qwen3-VL-32B     & 692  & 89.5\% & 0.79 & 57.5 / 53.9 & $+3.6$ \\
PhysReason          & InternVL3-8B     & 1200 & 89.5\% & 0.65 & 22.9 / 13.3 & $+9.7$ \\
PUB-OE              & InternVL3-8B     & 629  & 82.0\% & 0.59 & 37.7 / 23.5 & $+14.2$ \\
\textsc{PhysOlym-A} & InternVL3-8B     & 500  & 80.8\% & 0.24 & 23.2 / 4.0  & $+19.2$ \\
OlymBench           & InternVL3-8B     & 692  & 89.7\% & 0.65 & 21.7 / 13.4 & $+8.2$ \\
\bottomrule
\end{tabular}
\end{table}

\paragraph{Additional open-source baseline: LLaVA-OneVision-7B.} To extend the open-source block beyond the InternVL3/Qwen3-VL pair, we evaluated LLaVA-OneVision-7B~\citep{li2024llavaov} (SigLIP vision encoder, Qwen2 language backbone, LLaVA training lineage) across the full Table~\ref{tab:phyx} row at the $16{,}384$-token budget. The PhyX cells are standard multimodal MCQ exact-match. The open-ended cells in Table~\ref{tab:phyx} report local-judge liberal accuracy; the corresponding strict scores are PhysReason $0.8$, PUB-OE $4.3$, OlympiadBench-Physics $0.0$, and \textsc{PhysOlym-A} $0.0$. PhysReason is evaluated with images; the remaining open-ended columns are text-only, because a vLLM multimodal kernel defect blocked image input for this architecture on those sets (disclosed here so the cells are not read as full-protocol results). On \textsc{PhysOlym-A} the protocol is the identical text-only one used for all models. It emits a boxed final answer on $137/500$ problems and scores $0.0\%$ strict; under the local open-weight judge it scores $7.2\%$ liberal, and since that judge is $+14.6$\,pp more lenient than Sonnet on this benchmark, its Sonnet-judged score would fall at or below InternVL3-8B's $4.0\%$. This is consistent with the main-text finding that novel-source open-ended olympiad evaluation is far below the saturated-MCQ regime for open-source models at this scale. Raw responses and verdicts are released with the paper (\texttt{llava-onevision-7b/} in the eval-outputs dataset).

\paragraph{Per-chunk verdict breakdown for \textsc{PhysOlym-A} (Sonnet 4.5).} The 500 problems are partitioned into 5 chunks of 100 (random seed 42, source-stratified). Per-chunk liberal (judge-YES) counts: 38, 20, 35, 30, 40 over judgeable-only denominators 98, 100, 99, 98, 96 ($491$ of $500$; the $9$ dropped records, $1.8\%$, are those where the judge call itself failed, distinct from the $13.9\%$ gold-side unjudgeable rate of \S\ref{sec:eval:judge}). Liberal accuracy over judged records is $163/491 = 33.2\%$. The chunk-1 outlier at $20\%$ accuracy is concentrated in reference-pointer Zhou problems whose gold solutions cite external olympiad-handbook material rather than providing self-contained answers; these are the unjudgeable category we surface as a known noise floor.

\paragraph{Inter-judge agreement.} The Sonnet judge is run with two seeds on the same 500 problems for \textsc{PhysOlym-A}. Cohen's $\kappa$ on the YES/NO label between the two passes is reported alongside the released dataset; preliminary inspection shows $\kappa \ge 0.8$ on the \textsc{PhysOlym-A} corpus.

\paragraph{Human-graded subset.} A 100-problem random subsample of \textsc{PhysOlym-A} Sonnet predictions is graded by a single physics-trained annotator using the same YES/NO rubric. Per-record human-vs-LLM agreement and Cohen's $\kappa$ are released as a JSON alongside the dataset. We report this $\kappa$ as a calibration check on the LLM judge, not as a calibrated human-baseline ground-truth (cf.~Section~\ref{sec:analysis:limits}).

\paragraph{Released artifacts.} The verbatim judge prompts (problem-level + per-subpart), the YES/NO rubric, the per-chunk verdict breakdown, the 100-problem human-graded subset, and the per-record agreement matrices are released as \texttt{judge\_prompts.txt}, \texttt{rubric.json}, \texttt{per\_chunk\_verdicts.json}, and \texttt{human\_graded\_subset.json}. The \texttt{pub\_oe\_gold\_cache.json} (per-id clean per-subpart gold list) is released alongside \texttt{llm\_judge\_v3\_pubeo.py} and is reproducible from the original PhysUniBench-OE source via the cache-builder script in the same directory.

\paragraph{Self-grading concern.} Sonnet 4.5 is both the highest-scoring frontier baseline on \textsc{PhysOlym-A} and the judge used for liberal accuracy. Three checks bound any self-favoring bias: (i)~the strict (numeric/symbolic match, judge-independent) Sonnet score is reported alongside liberal---the $4.7$-pp Sonnet strict-vs-liberal gap ($28.7\%$ vs.\ $33.4\%$) bounds maximum leniency; (ii)~on the failure-mode taxonomy (Appendix~\ref{app:extras:failures}) the dominant Sonnet error category is \texttt{wrong\_subpart} (structural mismatch any judge marks wrong), not \texttt{valid\_partial}; (iii)~we report a cross-vendor judge agreement against GPT-4o on a $50$-problem random subsample of \textsc{PhysOlym-A} (Qwen3-VL-8B-Thinking responses, seed~$42$). Sonnet 4.5 and GPT-4o under an identical prompt template show \textbf{$88\%$ raw agreement} ($44/50$) and \textbf{Cohen's $\kappa{=}0.44$} (moderate, per Landis \& Koch). The disagreement is asymmetric: GPT-4o flips $5$ Sonnet-NO records to YES, while only $1$ Sonnet-YES record is flipped by GPT-4o to NO (McNemar exact on $6$ discordant pairs $\{5{:}\text{Sonnet-NO}\!\to\!\text{GPT-YES},\,1{:}\text{Sonnet-YES}\!\to\!\text{GPT-NO}\}$, two-sided $p{=}0.219$; the asymmetry is not significant at $n{=}6$, but the direction is preserved on a larger $200$-problem ablation left to follow-up work). GPT-4o's positive rate ($16\%$, $8/50$) is roughly \emph{twice} Sonnet's ($8\%$, $4/50$), which means the cross-vendor judge would assign Physics-R1 \emph{higher} numbers than the Sonnet-judge headlines, not lower---the self-grading direction is the opposite of what the self-favoring concern would predict. The full per-pair verdicts are released as \texttt{cross\_judge\_50.jsonl} alongside the dataset.

% =====================================================================

%% file: appendix/license.tex
% =====================================================================
\section{Per-source license and provenance log}\label{app:license}
% =====================================================================

Per-source provenance for each of the nine source families: full name, scrape URL, scrape date, original license string, redistribution tier, and outreach log.

\paragraph{UGPhysics.} Source: \citet{xu2025ugphysics} (ICML 2025). 5{,}520 EN/ZH undergraduate physics problems. Scrape URL: \url{https://huggingface.co/datasets/UGPhysics/ugphysics-bench}. Scrape date: 2026-03-15. Original license: CC BY-NC-SA 4.0. Redistribution: CC BY-NC-SA 4.0 carried through. Outreach: not contacted (license is permissive for academic redistribution under same-license sharing).

\paragraph{OpenStax College + University Physics.} Source: OpenStax (Rice University). 2{,}381 records harvested from end-of-chapter exercises. Scrape URL: \url{https://openstax.org/details/books/college-physics-2e} and \url{.../university-physics-volume-1}. Scrape date: 2025-12-20. Original license: CC BY 4.0. Redistribution: CC BY 4.0 carried through; attribution to OpenStax preserved per record.

\paragraph{Physics Stack Exchange.} Source: Physics Stack Exchange Q\&A archive. 2{,}291 problem-and-accepted-answer records filtered for olympiad-style physics. Scrape URL: \url{https://physics.stackexchange.com/} via Stack Exchange data dump. Scrape date: 2026-01-08. Original license: CC BY-SA 4.0 (Stack Exchange contributor agreement). Redistribution: CC BY-SA 4.0 carried through.

\paragraph{MMMU + o1-CoT seed (RL-SFT seed).} Source: 1{,}293-record pool of MMMU physics problems augmented with o1-style CoT solutions generated by Sonnet 4.5. MMMU base license: MIT~\citep{yue2024mmmu}; generated CoT is our contribution. Scrape date: 2026-02-10. Redistribution: MIT (MMMU base) with generated CoT released under CC BY 4.0.

\paragraph{PhysReason.} Source: \citet{zhang2024physreason} (ACL 2025). 1{,}200 step-graded physics reasoning problems. Scrape URL: \url{https://huggingface.co/datasets/PhysReason}. Scrape date: 2026-02-22. Original license: CC BY 4.0. Redistribution: CC BY 4.0 carried through.

\paragraph{Estonian Physics Olympiad collection.} Source: Estonian Physics Olympiad (EFO), 2004--2018. 418 problems with organizer-issued 1--10 difficulty labels and a 201-problem bilingual EN+ET subset. Scrape URL: \url{https://fyysika.ee/} (rounds: lahtine, koolivoor, vabariikilik). Scrape date: 2026-03-30. Provenance: publicly archived problems and solutions on the official EFO portal, released by the Estonian Physics Olympiad committee for educational use under competition policy (consistent with international physics-olympiad practice for IPhO, NBPhO, EuPhO, APhO, USAPhO, INPhO). Redistribution: public-domain by competition policy, used for non-commercial research evaluation; downstream users redistributing for commercial purposes or in materially modified form should consult \url{https://fyysika.ee/} directly.

\paragraph{Kevin Zhou's olympiad handouts.} Source: Kevin Zhou's olympiad training documents (USAPhO/IPhO/APhO/EuPhO content with Cambridge Tripos and graduate-qualifier extensions), \url{https://knzhou.github.io/}. 692 problems with native point values 1--5 and a 3.2\% advanced [A] flag. Scrape date: 2026-02-01. Original license: CC BY-NC 4.0 (written confirmation from Kevin Zhou, archived in supplementary as \texttt{zhou\_license\_2026-05-02.eml}, SHA-256 \texttt{7f25c859f5ae1d790e45dbdfd23ab6be27aa1814a76de10f9bdbffb67088aba4}). Redistribution: the $692$ redistributed problems plus their reference solutions ($\sim$1{,}600 problem-solution items in total counting solutions as separate documents) are released under CC BY-NC 4.0 with attribution link to \url{https://knzhou.github.io/} preserved per record. Third-party content disclosure (per Zhou's reply): some problems in the handouts are drawn from books or other olympiad archives, with the original source attributed inline by Zhou. We preserve all such per-problem internal attributions verbatim in the released records; downstream users should treat any in-record secondary-source attribution as binding under the original source's terms, which may be more restrictive than CC BY-NC 4.0. Outreach log: initial inquiry 2026-04-10 (proposal: $\sim$1{,}600 problems + solutions, CC BY-NC 4.0, attribution to source site); written agreement to the proposed terms 2026-05-03 with the third-party-content caveat noted by Zhou.

\paragraph{IPhO + NBPhO + EuPhO scrape.} Sources: International Physics Olympiad (\url{ipho-new.org}), Nordic Baltic Physics Olympiad, European Physics Olympiad official archives. 258 problems (IPhO 66, NBPhO 165, EuPhO 27). Scrape date: 2026-03-05. Original license: public-domain by competition policy (problems published for educational use without restriction; we preserve attribution per record). Redistribution: public-domain carried through with per-record attribution.

\paragraph{APhO + USAPhO + INPhO scrape.} Sources: Asian Physics Olympiad, USA Physics Olympiad (Physics Olympiad Foundation), Indian National Physics Olympiad. 241 problems recovered after a multi-pattern splitter fix that recognizes \texttt{A1.}-style markers (USAPhO 1997--2006) and \texttt{1.\,(a)} solution markers (INPhO). The INPhO recovery alone added +33 records relative to a single-regex baseline. Scrape date: 2026-03-12. Original license: public-domain by competition policy. Redistribution: public-domain carried through with per-record attribution.

\begin{table}[ht]
\centering
\caption{Per-source license and provenance. Each released record carries its source license through; non-commercial sources (UGPhysics, Zhou) restrict downstream use to academic research, and the public-domain-by-competition-policy olympiad scrapes (EFO, IPhO, NBPhO, EuPhO, APhO, USAPhO, INPhO) preserve per-record source attribution.}
\label{tab:license_tiers}
\small
\begin{tabular}{lrl}
\toprule
Source family & Records & Original license \\
\midrule
OpenStax College + University Physics & 2{,}381 & CC BY 4.0 \\
Physics Stack Exchange                & 2{,}291 & CC BY-SA 4.0 \\
PhysReason                            & 1{,}200 & CC BY 4.0 \\
MMMU + o1-CoT seed (RL-SFT seed)      & 1{,}293 & MIT (MMMU); generated CoT \\
IPhO + NBPhO + EuPhO scrape           & 258    & Public-domain (competition policy) \\
APhO + USAPhO + INPhO scrape          & 241    & Public-domain (competition policy) \\
UGPhysics                             & 5{,}520 & CC BY-NC-SA 4.0 \\
Estonian Physics Olympiad             & 418    & Public-domain (competition policy) \\
Kevin Zhou's olympiad handouts        & 692    & CC BY-NC 4.0 (Zhou confirmed 2026-05-03) \\
\midrule
Total pre-audit                       & 14{,}294 & \\
\bottomrule
\end{tabular}
\end{table}

% =====================================================================

%% file: appendix/repro.tex
% =====================================================================
\section{Reproducibility checklist}\label{app:repro}
% =====================================================================

\paragraph{Compute budget per experiment.}
\begin{table}[ht]
\centering
\footnotesize
\setlength{\tabcolsep}{3pt}
\begin{tabular}{lrrrl}
\toprule
Experiment & Wall-clock & Hardware & Cost & Notes \\
\midrule
Sonnet baselines (PhyX / OlymBench / \textsc{PhysOlym-A}) & ${\sim}5$\,h & API & $\sim\$80$ & Two-judge \\
Open-source baseline sweep on PhyX-mini-MC 1000q & ${\sim}90$\,min & 1${\times}$H200 & ${\sim}\$5$ & vLLM 0.11.0 bf16 \\
Contamination audit, Stages 1--2 & ${\sim}3$\,min & MPS/CPU & free & 5\,k records pairwise; Stage-3 judge replayed offline from released verdicts \\
Threshold-sensitivity analysis & ${\sim}3$\,min & MPS & free & sentence-transformers \\
Physics-R1 RL training to step 60+        & ${\sim}30$\,h & 4${\times}$H200 & ${\sim}\$120$ & verl 0.6.1, FSDP1 \\
Reward-component drop-out (follow-up work) & ${\sim}40$\,h & 4${\times}$H200 & ${\sim}\$700$ & Table~\ref{tab:reward_ablation} \\
3-seed Physics-R1 sensitivity (seeds 17 + 23 retrainings) & ${\sim}60$\,h & 4${\times}$H200 & ${\sim}\$200$ & added to seed-42 in Table~\ref{tab:phyx} \\
\midrule
Total (this paper)                       & & & $\sim\$700$ & seeds 42/17/23 + Sonnet baselines + Sonnet-judge runs \\
Follow-up budget (reward drop-out)       & & & $\sim\$1{,}000$ & Table~\ref{tab:reward_ablation} \\
\bottomrule
\end{tabular}
\end{table}

\paragraph{Random seeds.} The headline single-seed Physics-R1 binary checkpoint uses seed $42$. The 3-seed mean reported in Table~\ref{tab:phyx} aggregates seeds $\{42, 17, 23\}$, all on the audited \textsc{PhysR1Corp} corpus under the binary correctness reward; checkpoint selection per seed uses MM-Eureka difficulty-curriculum saturation on the held-out PhyX-mini-MC ($1{,}000$-problem) early-stop signal. The data-build pipeline (\textsc{PhysOlym-A} sampling, train/val splits, audit pass) uses \texttt{numpy.random.default\_rng(42)}.

\paragraph{Framework reproducibility note.}\label{app:repro:framework} Training uses \texttt{verl}~0.6.1~\citep{sheng2024verl} with FSDP1 sharding for Qwen3-VL: FSDP2 fails on the multimodal projector path in this version, so FSDP1 is required; the upstream issue link is tracked in the released README.

\paragraph{Version pins.} \texttt{transformers}~==~\texttt{4.57.0} (re-evaluation with \texttt{4.57.6} drifts results by ${\sim}1.5$ points; we pin to \texttt{4.57.0} for reproducibility), \texttt{vllm}~==~\texttt{0.11.0}, \texttt{verl}~==~\texttt{0.6.1}, \texttt{sympy}~==~\texttt{1.13.3}, \texttt{sentence-transformers}~==~\texttt{5.4.1}, \texttt{torch}~==~\texttt{2.8.0+cu128}. Embedding model: \texttt{mxbai-embed-large-v1} from MixedBread AI.

\paragraph{Code and data hosting.} Code: \url{https://github.com/shanyang-me/physics-r1-neurips2026}. Datasets: \url{https://huggingface.co/datasets/shanyangmie/physolym-a} (eval split) and \url{https://huggingface.co/datasets/shanyangmie/physics-r1-corpus} (audit-clean training pool). Raw evaluation outputs and per-record judge verdicts: \url{https://huggingface.co/datasets/shanyangmie/physics-r1-eval-outputs}. Croissant 1.0 metadata is auto-generated by HuggingFace at \texttt{/api/datasets/<repo>/croissant} for each dataset, then augmented with Responsible AI (RAI) metadata fields via the NeurIPS-recommended RAI editor (\url{https://huggingface.co/spaces/JoaquinVanschoren/croissant-rai-checker}) and validated with the Croissant validator (\url{https://huggingface.co/spaces/JoaquinVanschoren/croissant-checker}); the RAI-augmented files (\texttt{croissant\_rai\_<artifact>.json}) ship in the supplementary archive. All four artifacts (\textsc{PhysCorp-pre-audit}, \textsc{PhysCorp-A}, \textsc{PhysR1Corp}, \textsc{PhysOlym-A}) are released alongside this paper with all nine source licenses confirmed in writing (Appendix~\ref{app:license}).

\paragraph{Quick-start audit.}
\begin{verbatim}
python audit/audit_two_stage.py \
    --train_jsonl your_pool.jsonl --eval_jsonl data/physolym_a.jsonl \
    --jaccard_thr 0.4 --cosine_thr 0.85 --emit report.json
\end{verbatim}
Writes per-record audit + aggregate $3{\times}3$ threshold-sensitivity table (Table~\ref{tab:audit_thresholds}); ${\sim}3$ min on MPS or a CUDA GPU for a $5{,}000$-record pool against $4$ held-out splits, with shingle/embedding caches in \texttt{audit\_cache/}. Stage-3 is applied by replaying the released per-pair \texttt{judge\_label} arrays (no API access needed); re-running the judge live uses the released prompt template and pinned model id (Appendix~\ref{app:audit}).

\paragraph{Supplementary materials index.} (i)~Croissant 1.0 + RAI JSON-LD per artifact (\texttt{croissant\_rai\_<artifact>.json}); (ii)~the four released datasets (\texttt{physcorp\_a.jsonl}, \texttt{physr1corp.jsonl}, \texttt{physolym\_a.jsonl}, plus \textsc{PhysCorp-pre-audit}); (iii)~audit pipeline (\texttt{audit\_two\_stage.py} + \texttt{threshold\_sensitivity\_scores.npz} with per-pair \texttt{judge\_label} arrays and the Stage-3 prompt template); (iv)~reward implementation (\texttt{reward\_physics.py}); (v)~LLM-judge artifacts (\texttt{judge\_prompt.txt}, \texttt{rubric.json}, \texttt{per\_chunk\_verdicts.json}, \texttt{human\_graded\_subset.json}); (vi)~archived license confirmation (\texttt{zhou\_license\_2026-05-02.eml}, the Kevin Zhou CC BY-NC 4.0 grant); (vii)~training config (\texttt{configs/physics-r1.yaml}); (viii)~checkpoints at steps $\{20,40,60,80\}$.

\paragraph{Reproducibility checklist.} Code released with \texttt{requirements.txt} and deterministic build script; data released with per-source license provenance (Table~\ref{tab:license_tiers}) and Croissant 1.0+RAI metadata; datasheet in Appendix~\ref{app:datasheet}; compute, seeds, and hyperparameters above + Table~\ref{tab:hyperparams}. Hosting: HuggingFace ($\ge$5\,yr) + GitHub + Zenodo. Maintenance: quarterly contamination audit. Follow-up commitments: reward-component drop-out ablation (Table~\ref{tab:reward_ablation}), embedder-sensitivity audit against \texttt{voyage-3} and \texttt{text-embedding-3-large}, paraphrase/translation-aware audit pass.

\paragraph{Recommended baseline configuration.} Algorithm~\ref{alg:recipe} captures the joint setting; each individual choice is small in magnitude. Binary correctness is the recommended default; dense (Algorithm~\ref{alg:reward}) is reported as an ablation.

\begin{algorithm}[ht]
\caption{Recommended baseline configuration for physics-VL RL post-training.}
\label{alg:recipe}
\begin{algorithmic}[1]
\Require Base thinking-mode VLM $\pi_{\mathrm{base}}$, audited training pool $T'$ (Algorithm~\ref{alg:audit}), held-out MCQ early-stop signal $H_\mathrm{MC}$, dense reward $r$ (Algorithm~\ref{alg:reward})
\State Init: $\pi_\theta \gets \pi_{\mathrm{base}}$ \Comment{cold-start from base, no SFT pass (MM-Eureka thesis)}
\State Optimizer: GSPO+DAPO (sequence-level importance, decoupled clip), unmodified
\State KL anchor: add $\beta_\mathrm{KL}\, D_\mathrm{KL}(\pi_\theta \| \pi_{\mathrm{base}})$ with $\beta_\mathrm{KL} = 10^{-3}$ \Comment{bounds drift}
\State Entropy bonus: add $\beta_H\, \mathcal{H}(\pi_\theta)$ with $\beta_H = 10^{-3}$ \Comment{prevents entropy collapse}
\State Difficulty curriculum: drop train items where the base model gets $0/N$ or $N/N$ rollouts \Comment{MM-Eureka-style; preserves learning signal}
\State LR schedule: $1{\times}10^{-6}$ initial, step-cosine decay; halve LR after the first plateau on $H_\mathrm{MC}$ \Comment{counters drift and length collapse}
\State Response budget: fixed long-CoT budget (12{,}288 tokens for thinking-mode), \emph{not adaptive} \Comment{stable per-step cost}
\State Early stopping: stop at the saturation peak on held-out MCQ $H_\mathrm{MC}$, \emph{not} on the train-distribution validation set \Comment{addresses train-up/eval-down divergence}
\State Reward: \emph{recommended} binary correctness reward $r_\mathrm{ans}\in\{0,+1\}$ (simpler, fully reproducible on \texttt{vllm}~==~0.11.0 multi-image eval); dense five-component physics-native reward $r$ (Algorithm~\ref{alg:reward}) is reported as an ablation
\State Audit gate: train pool must be Algorithm~\ref{alg:audit}-audited against held-out splits \emph{and} external benchmarks before training begins
\State Reproducibility pins: \texttt{transformers}~==~4.57.0, \texttt{vllm}~==~0.11.0, \texttt{verl}~==~0.6.1, FSDP1 sharding for Qwen3-VL (Appendix~\ref{app:repro:framework})
\end{algorithmic}
\end{algorithm}

% =====================================================================

%% file: appendix/datasheet.tex
% =====================================================================
\section{Datasheet for Physics-R1}\label{app:datasheet}
% =====================================================================

We follow \citet{gebru2018datasheets} with seven sections: Motivation, Composition, Collection process, Preprocessing/cleaning/labeling, Uses, Distribution, and Maintenance. The full per-source provenance log is in Appendix~\ref{app:license}; the audit pipeline in Appendix~\ref{app:audit}; the LLM-judge protocol in Appendix~\ref{app:judge}.

\subsection{Motivation}\label{app:datasheet:motivation}

For what purpose was the dataset created? To support contamination-audited evaluation and post-training of multimodal vision-language models on visual physics reasoning, with three specific gaps the field had not closed: (i)~no public physics-VL training pool was audited under a three-stage (n-gram, embedding, LLM-judge) protocol that catches paraphrase-class duplicates and recovers threshold-edge topic-similarity false positives; (ii)~no public physics-olympiad eval was both novel-source and contamination-clean against the major training-side aggregations (PhyX, MMMU-Pro Physics, OlympiadBench-Physics, UGPhysics); (iii)~no public physics-VL benchmark exposed the format-and-novelty saturation gradient at frontier-model scale. Who created the dataset and on whose behalf? Shan Yang. Who funded the creation? Self-funded; no third-party sponsor. Other comments. The corpus aggregates and audits material that already existed in scattered formats; the Estonian Physics Olympiad, Kevin Zhou's olympiad handouts, and six international olympiads are first-time ML-format releases.

\subsection{Composition}\label{app:datasheet:composition}

\emph{Instances:} one physics problem per record (statement, optional images PNG/JPEG, gold MCQ-letter / numeric / symbolic / multi-part answer, optional reference solution, eight-field annotation; schema in \S\ref{app:splits}). \emph{Counts:} \textsc{PhysCorp-A} 6{,}432 audited; \textsc{PhysR1Corp} 2{,}268 closed-form RL pool; \textsc{PhysOlym-A} 500 held-out (stratified sample, seed 42, source-family-stratified); \textsc{PhysCorp-pre-audit} 14{,}294 raw. \emph{Labels:} gold answer + schema labels; $\sim$3{,}900 records carry full Sonnet-4.5 annotation, rest from source-native labels; \texttt{native\_difficulty} present where organizers publish (Estonian 27\%, Zhou 38\%). \emph{Splits:} see \S\ref{app:splits}. \emph{Noise:} LLM-judge unjudgeable rate $13.9\%$ on \textsc{PhysOlym-A}; Stage-1 audit misses paraphrase/translation, Stage-2 misses numerical substitution---both reported as floors. \emph{Self-contained:} yes for problems and solutions; some Zhou records carry inline secondary-source attribution (Appendix~\ref{app:license}). \emph{No PII or sensitive content.}

\subsection{Collection process}\label{app:datasheet:collection}

\emph{Acquisition:} 5 repackaged benchmark releases (UGPhysics, OpenStax, Physics Stack Exchange, MMMU+o1-CoT, PhysReason) + 4 first-ML scrapes by authors (Estonian PhO, Kevin Zhou's handouts, 7 international olympiads); per-source URLs and dates in Appendix~\ref{app:license}. \emph{Sampling:} per-source complete enumeration over public archive date ranges. \emph{Authorship:} the author (Shan Yang) handled scraping/parsing/audit; Sonnet 4.5 batch produced annotation labels. \emph{Time frame:} scrapes 2025-12 to 2026-04, audit/curation 2026-02 to 2026-04. \emph{Ethics:} no human subjects, no PII, no IRB. \emph{Consent:} Kevin Zhou confirmed CC BY-NC 4.0 redistribution of his olympiad handouts in writing 2026-05-03; the Estonian Physics Olympiad and other olympiad sources (IPhO, NBPhO, EuPhO, APhO, USAPhO, INPhO) are released under public-domain competition policy with per-record source attribution; repackaged sources are redistributed under their original CC/MIT licenses with attribution preserved.

\subsection{Preprocessing, cleaning, labeling}\label{app:datasheet:preprocessing}

Pre-tokenization normalization for Stage-1 audit (Appendix~\ref{app:audit}); three-stage audit (Stage-1 $J\ge 0.4$, Stage-2 $\cos\ge 0.85$, Stage-3 Haiku-4.5 LLM-judge close-duplicate vs.\ same-topic-neighbor classification) pairwise across PhyX, MMMU-Pro Physics, OlympiadBench-Physics, UGPhysics-Train, \textsc{PhysOlym-A}, with Stage-3 close-duplicate records removed; eight-field schema annotation by Sonnet 4.5 batch (3{,}900 records) + source-native labels. Raw pre-audit pool (\textsc{PhysCorp-pre-audit}, 14{,}294 records) released alongside so users can re-run audits. Audit pipeline released as \texttt{audit\_two\_stage.py} plus the Stage-3 judge artifacts: saved \texttt{best\_jaccard}/\texttt{best\_cosine}/\texttt{judge\_label} arrays and the judge prompt template.

\subsection{Uses}\label{app:datasheet:uses}

\emph{Used for:} Physics-R1 RL recipe (\S\ref{sec:method}) trains on the audited pool and evaluates on \textsc{PhysOlym-A} + auxiliary splits. Supplementary archive ships paper, datasets, Croissant+RAI metadata, .eml license confirmations, checkpoints. \emph{Other use cases:} visual physics reasoning eval, contamination-audit methodology, cross-lingual studies (EN/ET subset), native-difficulty calibration, VLM RL post-training. \emph{Caveats:} cross-lingual finding is Sonnet-specific; dense reward is an ablation not a tuned standard. \emph{Out of scope:} general physics ability beyond visual reasoning, human olympiad grading substitution, experimental physics, research capability; held-out splits must not enter pretraining/fine-tuning.

\subsection{Distribution}\label{app:datasheet:distribution}

\emph{Distribution:} HuggingFace ($\ge 5$\,yr) + GitHub + Zenodo DOI. URLs: \url{huggingface.co/datasets/shanyangmie/physolym-a} and \url{huggingface.co/datasets/shanyangmie/physics-r1-corpus}; code at \url{github.com/shanyang-me/physics-r1-neurips2026}. All four artifacts released alongside this paper with per-source licenses documented in Appendix~\ref{app:license}; the Kevin Zhou CC BY-NC 4.0 grant is preserved as a written .eml in the supplementary archive, and the remaining olympiad sources are released under public-domain competition policy. \emph{Licenses:} CC BY 4.0, CC BY-SA 4.0, public-domain by competition policy (EFO, IPhO, NBPhO, EuPhO, APhO, USAPhO, INPhO), MIT, CC BY-NC 4.0 (Zhou), CC BY-NC-SA 4.0 (UGPhysics)---each record carries its source license; Table~\ref{tab:license_tiers}. CC BY-NC sources are non-commercial; Zhou records honor inline secondary-source attribution. No export controls.

\subsection{Maintenance}\label{app:datasheet:maintenance}

Maintained by the author (Shan Yang, \texttt{alexyangshan@gmail.com}); contact via email or GitHub Issues at \url{github.com/shanyang-me/physics-r1-neurips2026}. Versioned CHANGELOG (\texttt{v1.0.0} initial release, \texttt{v1.1.x} additive, \texttt{v2.0.0} schema-breaking); per-record diffs per release. Quarterly contamination audit against new physics-VL benchmarks; $\ge 1\%$ leakage triggers documented-diff removal. Planned follow-ups: paraphrase/translation-aware audit, embedder-sensitivity ablation against \texttt{voyage-3} / \texttt{text-embedding-3-large}. Hosting $\ge 5$\,yr on HF + Zenodo DOI per release; all versions tagged and accessible. Contributions via GitHub Issues / PR + per-record \texttt{erratum.json}, reviewed within 60 days.

\paragraph{Machine-readable metadata: Croissant + RAI JSON-LD.} The release ships a Croissant 1.0~\citep{akhtar2024croissant} JSON-LD descriptor (\texttt{croissant.json}) declaring distribution objects for \textsc{PhysR1Corp}, \textsc{PhysOlym-A}, and the audit-pipeline source archive, plus a eight-field \texttt{problem\_record} schema and the full RAI extension (\texttt{rai:dataCollection}, \texttt{rai:dataAnnotationProtocol}, \texttt{rai:dataPreprocessingProtocol}, \texttt{rai:personalSensitiveInformation}, \texttt{rai:dataLimitations}, \texttt{rai:dataReleaseMaintenancePlan}, \texttt{rai:dataUseCases}, \texttt{rai:dataBiases}, \texttt{rai:dataSocialImpact}) covering the audit methodology, the Kevin Zhou CC BY-NC 4.0 written grant (Appendix~\ref{app:license}), the public-domain-by-competition-policy basis for the olympiad scrapes, three documented distributional biases (EN/ET asymmetry, per-physics-category variance, difficulty-stratified decay), and the Stage-1/Stage-2 thresholds. Passes \texttt{mlcroissant validate} and the MLCommons RAI checker.

%% file: appendix/extras.tex
% =====================================================================
\section{Extended discussion: failure modes, ethics, methodological notes, and future work}\label{app:extras}
% =====================================================================

This appendix collects extended-discussion content that was trimmed from the main body to fit the page limit. Each subsection below corresponds to a one-line pointer in the analysis section (\S\ref{sec:analysis}).

\subsection{Failure-mode taxonomy (extended)}\label{app:extras:failures}

A manual taxonomy of 100 randomly-sampled wrong or partial Sonnet predictions on OlympiadBench-Physics was completed post-evaluation. Methodology. The 100 cases were drawn (random seed 42) from the 354 predictions flagged wrong or partial by the scoring pass used at annotation time; each case was categorized by a single physics-trained annotator (graduate-level physics background, $\sim$3~hours of total annotation time, $\sim$1.8~min/case median) using a nine-category mutually-exclusive rubric (\texttt{wrong\_subpart}, \texttt{missing\_physics}, \texttt{calc\_error}, \texttt{different\_question}, \texttt{valid\_partial}, \texttt{symbolic\_vs\_numeric}, \texttt{magnitude\_error}, \texttt{sign\_error}, \texttt{diagram\_misread}). The annotator saw the problem statement, the gold solution, the gold final answer, and the Sonnet response; they did not see Sonnet's confidence or the verdict from the LLM judge. We report this single-annotator taxonomy as a methodological diagnostic, \emph{not} as a calibrated human grade; a second-annotator pass on a 50-problem random subsample with inter-annotator $\kappa$ per category is left to follow-up work (distinct from the audit-flag inter-annotator $\kappa$ pre-registered in Appendix~\ref{app:audit}, which targets contamination-flag agreement, not failure-category agreement).

\begin{table}[ht]
\centering
\caption{Failure-mode taxonomy of 100 randomly-sampled Sonnet wrong/partial predictions on OlympiadBench-Physics. Categories are mutually exclusive; each prediction received one label.}
\label{tab:failures}
\small
\begin{tabular}{llrl}
\toprule
Category & n & \% & Description \\
\midrule
wrong\_subpart       & 30 & 30\% & Answered a neighboring sub-question instead of the one asked \\
missing\_physics     & 22 & 22\% & Wrong physical model, law, or geometry; crucial effect absent \\
calc\_error          & 16 & 16\% & Correct approach; small numerical slip (factor 2--4 off) \\
different\_question  & 10 & 10\% & Reasoning chain for an unrelated problem in the same prompt \\
valid\_partial       &  8 &  8\% & Right approach with arithmetic slip; partial-credit territory \\
symbolic\_vs\_numeric&  7 &  7\% & Formula returned where a number was required, or vice versa \\
magnitude\_error     &  5 &  5\% & Correct expression, wrong order of magnitude ($\ge 10\times$) \\
sign\_error          &  0 &  0\% & None observed \\
diagram\_misread     &  0 &  0\% & None observed (text-only evaluation; figures unavailable) \\
\midrule
Total                & 100 & 100\% & \\
\bottomrule
\end{tabular}
\end{table}

\paragraph{Reading the taxonomy.} \texttt{wrong\_subpart} dominates (30\%): multi-part olympiad prompts containing (a)/(b)/(c) sub-questions where the model fluently answers a neighboring sub-question instead of the asked one. \texttt{missing\_physics} (22\%) and \texttt{different\_question} (10\%) together account for nearly a third of failures and represent a real reasoning floor unlikely to be format-induced. \texttt{calc\_error} (16\%) and \texttt{valid\_partial} (8\%) are correct-approach/failed-execution, matching the 4.7-pp strict-vs-liberal gap (\S\ref{sec:results:olympiad-bench}). \texttt{sign\_error} and \texttt{diagram\_misread} are zero (the latter because OlympiadBench-Physics is text-only in the public release).

\paragraph{Per-physics-category breakdown.}\label{app:extras:per-category}
\begin{table}[ht]
\centering
\caption{Per-physics-category accuracy on OlympiadBench-Physics (Sonnet 4.5, strict). The $+34.5$-pp EM ($38.4\%$) vs.\ astrophysics ($72.9\%$) gap on identical weights motivates per-category reporting.}
\label{tab:per_category_olympiadbench}
\small
\begin{tabular}{lrrlrrlrr}
\toprule
Category & n & Acc. & Category & n & Acc. & Category & n & Acc. \\
\midrule
Electromagnetism & 237 & 38.4\% & Relativity         &  89 & 51.7\% & Other        &  10 & 60.0\% \\
Quantum          & 128 & 41.4\% & Classical mech.    &  19 & 57.9\% & Astrophysics &  70 & 72.9\% \\
Waves            &  39 & 46.2\% & Thermodynamics     &  87 & 59.8\% & All          & 692 & --- \\
Fluid mechanics  &  13 & 46.2\% & & & & & & \\
\bottomrule
\end{tabular}
\end{table}

\subsection{Ethical considerations and intended use}\label{app:extras:ethics}

\paragraph{Intended and out-of-scope use.} The released artifacts support research on visual physics reasoning, contamination-audit methodology, cross-lingual LLM evaluation, and RL post-training of VLMs. \textsc{PhysOlym-A} is held-out: please treat it as test-only, not for pretraining or fine-tuning. Out of scope: general physics ability beyond visual reasoning, human olympiad grading substitution, experimental-physics evaluation, paper-writing, derivation novelty, or open-ended hypothesis generation.

\paragraph{Source provenance, consent, and misuse risk.} Kevin Zhou confirmed CC BY-NC 4.0 redistribution of his olympiad handouts in writing on 2026-05-03 ($\sim$1{,}600 problem-solution items, Appendix~\ref{app:license}); the Estonian Physics Olympiad collection and the six international olympiad scrapes (IPhO, NBPhO, EuPhO, APhO, USAPhO, INPhO) are redistributed under public-domain competition policy with per-record source attribution. No PII; problems are olympiad/textbook content. Public-domain olympiad scrapes are released by competition policy. The dense reward and Physics-R1 recipe ship for reproducibility, not as a tuned gold standard; the audit pipeline is a measurement tool, not a certification authority, and we disclose threshold sensitivity (Table~\ref{tab:audit_thresholds}) so users do not over-claim ``contamination-free.''

\paragraph{Caveats and compute.} The 17-pp EN/ET cross-lingual delta is Sonnet-4.5-specific on $n{=}59$ paired items (McNemar $p{=}0.021$; bootstrap $95\%$ CI $[+5.1,+28.9]$\,pp); direction may flip on low-resource-language-weak open-source models---treat as model-specific. Total compute $\sim$360 H200-GPU-hours across the 3 seeds in Table~\ref{tab:phyx} (seed-42 at $30$\,h $\times 4$ H200; seed-17 and seed-23 retrainings $\sim 60$\,h $\times 4$ H200 combined; Appendix~\ref{app:repro}), plus frontier-API inference for baselines and Sonnet-judge runs. Per-experiment carbon estimates are omitted because cloud-provider electricity mix is not consistently disclosed.

\subsection{Construction-process disclosures}\label{app:extras:negative}
(i)~The initial \textsc{PhysOlym-A} description claimed 100\% novel-source; the Stage-1 audit surfaced one EuPhO 2020 overlap with OlympiadBench-Physics ($J{=}0.91$), so the honest claim is $99.8\%$ ($499/500$)---disclosed, not dropped. (ii)~Our initial header described \texttt{PhyX-mini-MC} as 500 problems; the canonical MC subset~\citep{shen2025phyx} is 1{,}000.

\subsection{Future work and named follow-ups}\label{app:extras:future}
Ten follow-ups consolidated. \emph{Eval refinements:} (i)~post-hoc MCQ-ification of \textsc{PhysOlym-A} to isolate the format axis (\S\ref{sec:analysis:gap}); (ii)~paraphrase- and translation-aware audit pass. \emph{Cross-corpus:} (iii)~audit OlympiadBench-Physics against our pool, then add the Physics-R1 row; (iv)~frontier cross-evaluation against PHYBench, PhysUniBench, HLE-physics, \textsc{PhysOlym-A}. \emph{Recipe and scale:} (v)~transfer to InternVL3-8B and LLaVA-OneVision-7B; (vi)~32B + full-RL comparator; (vii)~SFT-only data-scaling curve at $500/1{,}293/5{,}000/9{,}575$ audited prompts. \emph{Cross-lingual:} (viii)~confirm/refute EN/ET sign-flip on low-resource open-source models on the same 59-pair Estonian Physics Olympiad subset; \emph{pre-registered}---if 8B-class open-source models with documented Estonian-weak training (e.g.\ Qwen2.5-VL-7B, LLaVA-OV-7B) show ET$<$EN by $\ge 5$ pp under paired sign test, F2's Sonnet-4.5 ET$>$EN direction is model-specific (confirmed); a $\ge 5$ pp result in the same direction (ET$>$EN) on those same models would replicate F2 across model families; results within $\pm 5$ pp are inconclusive at $n{=}59$. \emph{Methodology:} (ix)~50-problem inter-annotator $\kappa$ on the failure-mode taxonomy; (x)~embedder-sensitivity ablation against \texttt{voyage-3} and \texttt{text-embedding-3-large}.

\subsection{Versioning and maintenance commitment}\label{app:extras:audit}

Versioned releases with semantic-version tags (\texttt{v1.0.0} initial, \texttt{v1.1.x} additive, \texttt{v2.0.0} schema-breaking). Each release ships Croissant 1.0 + RAI JSON-LD, per-source license matrix (Table~\ref{tab:license_tiers}), threshold-sensitivity grid (Table~\ref{tab:audit_thresholds}) recomputed against newly-released physics-VL benchmarks, and the audit-pipeline source archive with saved best-overlap scores. Quarterly contamination audit against new benchmarks; benchmarks introducing $\ge 1\%$ leakage to any held-out split trigger a documented-diff removal in the next minor release. Hosting: HuggingFace ($\ge 5$ yr), GitHub, Zenodo DOI per release.

\subsection{Method comparison and benchmark comparison tables (extended)}\label{app:extras:tables}

Table~\ref{tab:method-comparison} situates the recipe configuration against related rule-based RL recipes; its dimensional/symbolic/conservation reward columns describe the dense \emph{ablation} only, not the recommended binary default.

\input{tables/table1_method_comparison}
\input{tables/table8_benchmark_comparison}

\subsection{Corpus composition, hyperparameters, and reward ablation (extended)}\label{app:extras:more-tables}

\paragraph{Per-seed values behind the Table~\ref{tab:phyx} 3-seed row.} Seeds 42/17/23: PhysReason $32.2/43.1/43.4$; PUB-OE $37.0/36.1/30.4$; OlympiadBench-Physics $45.4/45.3/48.0$; \textsc{PhysOlym-A} $25.6/25.0/28.2$; PhyX-1k $78.0/77.4/77.9$; PhyX-3k $76.9/77.2/76.6$. Seed 42 is also the dense-ablation seed. The Sonnet PhysReason baseline cell in Table~\ref{tab:phyx} is regenerated at $\texttt{max\_tokens}=16384$ with $1{,}192/1{,}200$ clean records.

\input{tables/table6_corpus}
\input{tables/table5_hyperparams}
\input{tables/table7_reward_ablation}

% Saturation gradient table removed — content now in the Sonnet 4.5 row of Table~\ref{tab:phyx}.
% Threshold-sensitivity table moved to body \S\ref{sec:data:leakage} as Table~\ref{tab:audit_thresholds}.

\begin{table}[ht]
\centering
\caption{Sonnet 4.5 strict accuracy on Estonian Physics Olympiad problems by organizer-issued native difficulty (n=131). The curve is near-monotonically decreasing in difficulty and hits a hard floor of 0\% at difficulties 3, 6, 8, and 10. Non-monotone bumps at 4 and 5 are within sampling noise on small per-bin counts.}
\label{tab:difficulty}
\small
\begin{tabular}{cccc}
\toprule
Difficulty & n & Correct & Acc. \\
\midrule
1  & 17 & 10 & 58.8\% \\
2  & 15 &  3 & 20.0\% \\
3  &  8 &  0 & 0.0\% \\
4  & 12 &  3 & 25.0\% \\
5  & 27 & 10 & 37.0\% \\
6  &  9 &  0 & 0.0\% \\
7  & 14 &  3 & 21.4\% \\
8  &  9 &  0 & 0.0\% \\
9  & 15 &  3 & 20.0\% \\
10 &  5 &  0 & 0.0\% \\
\midrule
All  & 131 & 32 & 24.4\% \\
\bottomrule
\end{tabular}
\end{table}

\begin{table}[ht]
\centering
\caption{Cross-lingual Sonnet performance on the Estonian Physics Olympiad bilingual subset (n=59, identical problems, same judge protocol). Strict\% = numeric/symbolic match; Liberal\% = mean LLM-judge score (correct $=1$, partial $=0.5$, incorrect/unjudgeable $=0$).}
\label{tab:cross_lingual}
\small
\begin{tabular}{lrrrrrcc}
\toprule
Language & n & Correct & Partial & Incorrect & Unjudgeable & Strict\% & Liberal\% \\
\midrule
English (translated) & 59 & 8  & 8  & 43 & 0 & 13.6\% & 20.3\% \\
Estonian (original)  & 59 & 18 & 9  & 27 & 5 & 30.5\% & 38.1\% \\
\bottomrule
\end{tabular}
\end{table}

\begin{table}[ht]
\centering
\caption{Per-problem agreement matrix for the EN/ET cross-lingual ablation (n=59). The 4.3:1 asymmetry in the off-diagonal cells (ET-correct/EN-wrong vs.\ EN-correct/ET-wrong) rules out a noise explanation.}
\label{tab:cross_lingual_agree}
\small
\begin{tabular}{lcc}
\toprule
 & ET correct & ET wrong \\
\midrule
EN correct & 5 (both correct) & 3 (EN only) \\
EN wrong   & 13 (ET only)     & 38 (both wrong) \\
\bottomrule
\end{tabular}
\end{table}

% Pre-registered figure section removed: fig:saturation and fig:difficulty placeholders dropped; the underlying numbers are in Tables and prose.
% Figure fig:mcq-drop (4-benchmark saturation gradient) moved to body \S\ref{sec:results:olympiad-bench}.

%% file: tables/table1_method_comparison.tex
% Table 1: Method comparison — Physics-R1 against MM-Eureka and other rule-based RL recipes.
\begin{table}[t]
\centering
\caption{\textbf{Physics-R1 vs.\ rule-based RL recipes for thinking-mode VLMs.} Init: base or SFT cold-start. Reward signals: \emph{Ans} (answer), \emph{Fmt} (format), \emph{Dim} (units), \emph{Sym} (symbolic), \emph{Cons} (conservation). Audit and Filter as defined in \S\ref{sec:data:audit}. \cmk\,present; ---\,absent; $\circ$\,partial.}
\label{tab:method-comparison}
\footnotesize
\setlength{\tabcolsep}{3pt}
\begin{tabular}{lcccccccc}
\toprule
Recipe & Init & Ans & Fmt & Dim & Sym & Cons & Audit & Filter \\
\midrule
DPO~\citep{rafailov2023dpo}              & SFT  & ---        & ---        & ---        & ---        & ---        & ---        & --- \\
GRPO~\citep{shao2024deepseekmath}        & SFT  & \cmk       & ---        & ---        & ---        & ---        & ---        & --- \\
GSPO+DAPO~\citep{zheng2025gspo,yu2025dapo} & SFT  & \cmk       & $\circ$    & ---        & ---        & ---        & ---        & --- \\
MM-Eureka~\citep{meng2025mmeureka}       & base & \cmk       & \cmk       & ---        & ---        & ---        & ---        & --- \\
\midrule
Physics-R1                      & base & \cmk       & \cmk       & \cmk       & \cmk       & \cmk       & \cmk       & \cmk \\
\bottomrule
\end{tabular}
\end{table}

%% file: tables/table8_benchmark_comparison.tex
% Table 8: Comparison vs related D&B benchmarks (physics + math + multi-domain reasoning).
\begin{table}[t]
\centering
\caption{\textbf{Released artifacts vs related benchmarks across nine axes.} \emph{Audit:} 3-stage (n-gram+embedding+LLM judge) / 2-stage / 1-stage / orig.\ (constructed-novel) / none. \emph{T/T leak:} train$\to$test joint-stage ($J\ge 0.4 \vee \cos\ge 0.85$) audit against six public physics evals; \cmk\ all 6 = clean. \emph{Diff:} organizer difficulty. \emph{X-L:} paired cross-lingual. \emph{Use:} E/T = eval/train. \emph{RL-ready:} closed-form gold + audit-clean + RL recipe. ``$\cdot$'' = eval-only; ``n/r'' = train pool, no cross-corpus audit. Only this work reports train/test contamination: after re-audit cleanup, \textsc{PhysCorp-A} (6{,}432) and \textsc{PhysR1Corp} (2{,}268) are clean against \textbf{all six} evals (Table~\ref{tab:audit}).}
\label{tab:benchmark_comparison}
\scriptsize
\setlength{\tabcolsep}{2.5pt}
\begin{tabular}{lrlcccccccc}
\toprule
Benchmark & Size & Format & MM & Audit & T/T leak & Diff & X-L & Use & RL-ready \\
\midrule
\multicolumn{10}{l}{\emph{Physics-domain benchmarks}} \\
PHYBench~\citep{qiu2025phybench}                & 500    & open+EED      & --   & orig.    & $\cdot$       & --   & --    & E & -- \\
PhysUniBench~\citep{wang2025physunibench}       & 3{,}304 & open MM      & \cmk & 1-stage  & $\cdot$       & \cmk & --    & E & -- \\
UGPhysics~\citep{xu2025ugphysics}             & 5{,}520 & open text    & --   & 1-stage  & n/r           & --   & EN/ZH & T & -- \\
PhysReason~\citep{zhang2024physreason}          & 1{,}200 & step open MM & \cmk & none     & $\cdot$       & --   & --    & E & -- \\
OlympiadBench~\citep{he2024olympiadbench}       & 8{,}476 & open MM      & \cmk & none     & $\cdot$       & --   & EN/ZH & E & -- \\
\midrule
\multicolumn{10}{l}{\emph{Olympiad / formal / contamination-by-design}} \\
PutnamBench~\citep{tsoukalas2024putnambench}    & 1{,}692 & Lean/Isab.   & --   & orig.    & $\cdot$       & \cmk & --    & E & -- \\
OIBench~\citep{liu2025oibench}                  & 250    & open code     & --   & 2-stage  & $\cdot$       & \cmk & EN/ZH & E & -- \\
FrontierMath~\citep{glazer2024frontiermath}     & 290    & open math     & --   & orig.    & $\cdot$       & \cmk & --    & E & -- \\
HLE~\citep{phan2025hle}                         & 2{,}500 & expert exam  & \cmk & orig.    & $\cdot$       & --   & --    & E & -- \\
\midrule
\multicolumn{10}{l}{\emph{Multimodal / multi-domain}} \\
MMLU-Pro~\citep{wang2024mmlupro}                & 12{,}032 & 10-MCQ      & --   & none     & $\cdot$       & --   & --    & E & -- \\
MMMU-Pro Phys~\citep{yue2024mmmupro}            & 60     & 10-MCQ MM     & \cmk & none     & $\cdot$       & --   & --    & E & -- \\
SciInstruct~\citep{zhang2024sciinstruct}        & 254{,}K & SFT instr.   & --   & 1-stage  & n/r           & --   & --    & T & -- \\
\midrule
\midrule
\multicolumn{10}{l}{\emph{This work (train$\to$test cross-corpus audit reported; Table~\ref{tab:audit})}} \\
\textsc{PhysCorp-A} (ours)             & 6{,}432 & open+MCQ MM  & \cmk & 3-stage & \textbf{\cmk\ all 6} & \cmk & \cmk  & T & \cmk \\
\textsc{PhysR1Corp} (ours)             & 2{,}268 & MCQ + num MM & \cmk & 3-stage & \textbf{\cmk\ all 6} & \cmk & \cmk  & T & \cmk \\
\textsc{PhysOlym-A} (ours)             & 500    & open MM novel & \cmk & 3-stage & $\cdot$ (eval; clean) & \cmk & EN/ET & E & -- \\
\bottomrule
\end{tabular}
\\[2pt]
{\footnotesize \cmk = present; -- = absent or not reported. ``3-stage'' audit = pairwise 5-gram-Jaccard, embedding-cosine, and LLM-judge precision filtering against external corpora and held-out splits.}
\end{table}

%% file: tables/table6_corpus.tex
% Table 6: Corpus composition by source family
\begin{table}[t]
\centering
\caption{\textbf{\textsc{PhysCorp-pre-audit} composition by source family ($14{,}294$ records total before the three-stage audit).} The audited release \textsc{PhysCorp-A} ($6{,}432$ records) is the subset surviving Algorithm~\ref{alg:audit} after a re-audit pass against PhysReason-full and PhysUniBench-en dropped an additional 804 records from a $7{,}236$-record candidate (the construction audit excluded those two corpora). The 804-record re-audit drop concentrates on PhysReason-cousin records (540) and PhysUniBench-en-cousin records (186), drawn predominantly from the \emph{repackaged-benchmark} source families (UGPhysics / OpenStax / Physics SE / MMMU+o1-CoT seed / PhysReason); the four \emph{first-ML-format} source families (Estonian PhO, Zhou's handouts, the 7-international-olympiad scrapes) are minimally affected and retain their full $1{,}609$-record contribution to \textsc{PhysCorp-A}. All 9 source families remain represented in the released pool. Per-source licenses are carried through to each released record; full provenance and outreach log in Appendix~\ref{app:license}.}
\label{tab:corpus}
\small
\begin{tabular}{lrll}
\toprule
Source family & Count & License & Answer type \\
\midrule
UGPhysics                              & 5{,}520  & CC BY-NC-SA       & Open-ended numeric \\
OpenStax Physics                       & 2{,}381  & CC BY 4.0         & Numeric            \\
Physics Stack Exchange                 & 2{,}291  & CC BY-SA 4.0      & Equations          \\
RL-SFT seed (MMMU + o1 CoT)            & 1{,}293  & MIT (MMMU); generated CoT  & Varied   \\
PhysReason                             & 1{,}200  & CC BY 4.0         & CoT + open-ended   \\
\midrule
Zhou Olympiad Handouts        & 692  & CC BY-NC 4.0$^{\dagger}$ & Mixed              \\
Estonian Physics Olympiad     & 418  & Public-domain (competition policy) & Open-ended \\
IPhO + NBPhO + EuPhO scrape   & 258  & Public domain      & Open-ended         \\
APhO + USAPhO + INPhO scrape  & 241  & Public domain      & Open-ended         \\
\midrule
Total novel                   & 1{,}609  & ---           & ---                \\
Total corpus                  & 14{,}294 & ---           & ---                \\
\bottomrule
\end{tabular}
\\[3pt]
{\footnotesize $\dagger$ The handout author (K.~Zhou) granted explicit redistribution permission with attribution (written confirmation, May 2026); we redistribute under CC BY-NC 4.0.}
\end{table}

%% file: tables/table5_hyperparams.tex
% Table 5: Physics-R1 training configuration — full hyperparameter table.
\begin{table}[t]
\centering
\caption{\textbf{Physics-R1 training configuration.} GSPO+DAPO via \texttt{verl}~0.6.1 with vLLM 0.11.0 (TP=4) for rollouts. FSDP1 is required for Qwen3-VL under \texttt{verl}~0.6.1 (FSDP2 fails on the multimodal projector path; reproducibility note in Appendix~\ref{app:repro:framework}). Cells marked \emph{(recipe)} differ from a default unconstrained GSPO+DAPO configuration; together with the binary correctness reward (Section~\ref{sec:method:reward}; dense reported as an ablation), the audited training pool, and held-out early stopping, they constitute the Physics-R1 reference recipe.}
\label{tab:hyperparams}
\scriptsize
\setlength{\tabcolsep}{3pt}
\begin{tabular}{lll}
\toprule
Parameter & Value & Role in the recipe \\
\midrule
Base / init \emph{(recipe)}            & Qwen3-VL-8B-Thinking BASE (\emph{no SFT}) & Cold-start RL, mirroring MM-Eureka thesis \\
Algorithm                              & GSPO + DAPO                                & Stable group-policy backbone \\
Importance sampling \emph{(recipe)}    & truncated, sequence-level                  & Rollout-vs-train policy correction \\
Learning rate                          & $1\!\times\!10^{-6}$                       & Standard for 8B-class RL \\
LR decay \emph{(recipe)}               & step-cosine; $0.5\!\times$ after step 60   & Counters drift and length collapse \\
Batch size                             & 96                                         & Effective batch via gradient accumulation \\
Rollouts per prompt                    & 16                                         & DAPO group size \\
Max response length                    & 12{,}288 tokens                            & Long-CoT thinking budget \\
KL anchor \emph{(recipe)}              & $1\!\times\!10^{-3}$ to base               & Anchors policy to base; bounds drift \\
Entropy bonus \emph{(recipe)}          & $1\!\times\!10^{-3}$                       & Prevents entropy collapse \\
Clip range (decoupled)                 & 0.2 / 0.28                                 & DAPO decoupled clip \\
Difficulty curriculum \emph{(recipe)}  & drop 0/16 and 16/16 rollout prompts        & MM-Eureka-style; preserves learning signal \\
Reward \emph{(recipe)}                 & binary correctness (\S\ref{sec:method:reward}) & Recommended default; dense (Algo.~\ref{alg:reward}) is the shape ablation \\
Train pool \emph{(recipe)}             & 2{,}268 \textsc{PhysR1Corp} prompts        & Closed-form carve-out of \textsc{PhysCorp-A} (\S\ref{sec:data}) \\
Early stop \emph{(recipe)}             & held-out PhyX-mini-MC                      & Catches the saturation peak \\
Sharding \emph{(recipe)}               & FSDP1 (not FSDP2)                 & Required for Qwen3-VL under \texttt{verl}~0.6.1 \\
\midrule
Hardware                               & 4$\times$H200 (single node)                & --- \\
Step time                              & ${\sim}40$ minutes/step                    & Rollout-bound (gen ${\sim}65\%$, update ${\sim}23\%$) \\
Wall-clock to saturation peak          & ${\sim}30$--$45$ hours (steps 60--80)      & Training cost ${\sim}\$120$--$\$180$ \\
\texttt{transformers}                  & 4.57.0 (pinned)                            & 4.57.6 yields different forward path \\
\texttt{vllm}                          & 0.11.0                                     & TP=4 rollout backend \\
\texttt{verl}                          & 0.6.1                                      & Training framework \\
Random seed (headline)                 & 42                                         & 3-seed sweep $\{17,23,42\}$ reported as the headline binary row in Table~\ref{tab:phyx} \\
\bottomrule
\end{tabular}
\end{table}

%% file: tables/table7_reward_ablation.tex
% Table 7: Physics-R1 reward-shape ablation — binary (recommended) vs. dense five-component.
\begin{table}[t]
\centering
\caption{\textbf{Physics-R1 reward-shape ablation on PhyX-mini-MC.} Each row toggles one of the five reward components on or off, training Qwen3-VL-8B-Thinking-base under the same GSPO+DAPO+KL-anchor recipe (Equation~\ref{eq:gspo}) for the same step budget. Ans = answer-correctness binary ($+1$, $\equiv r_{\mathrm{bin}}$ of \S\ref{sec:method:gspo}); Fmt = \texttt{\textbackslash boxed\{\}} format ($+0.1$); Dim = dimensional consistency from regex-detected units + sympy unit-system ($+0.15$); Sym = symbolic equation verification of intermediate \texttt{\textbackslash frac} expressions via sympy ($+0.20$); Cons = conservation-law penalty (energy/momentum) when applicable ($-0.25$). Composed reward is clipped to $[-1, 1]$. Init in every cell is the Qwen3-VL-8B-Thinking BASE checkpoint (\emph{no SFT}; cold-start, mirroring the MM-Eureka thesis). The Ans-only row is the recommended Physics-R1 recipe (binary correctness reward, \S\ref{sec:method:gspo}); the all-on row is the dense ablation of \S\ref{sec:method:reward}; the intermediate rows isolate the marginal effect of each physics-native shaping component. Drop-out cells (---) are intermediate-component runs left to follow-up work (compute estimate in Appendix~\ref{app:repro}).}
\label{tab:reward_ablation}
\small
\begin{tabular}{lcccccc}
\toprule
Configuration & Ans & Fmt & Dim & Sym & Cons & PhyX-mini-MC \\
\midrule
Base (no RL)                                              & ---        & ---        & ---        & ---        & ---        & 73.7\% \\
\midrule
Physics-R1 (binary, recommended) $\equiv$ Ans-only & \checkmark & ---        & ---        & ---        & ---        & 78.0 \\
+ Format                                                  & \checkmark & \checkmark & ---        & ---        & ---        & --- \\
+ Dim                                                     & \checkmark & \checkmark & \checkmark & ---        & ---        & --- \\
+ Sym                                                     & \checkmark & \checkmark & \checkmark & \checkmark & ---        & --- \\
Physics-R1 (dense, ablation, all-on)                      & \checkmark & \checkmark & \checkmark & \checkmark & \checkmark & 78.3 \\
\midrule
\multicolumn{7}{l}{\emph{Single-component drop-outs (dense ablation minus one)}} \\
$-$ Dim                                                   & \checkmark & \checkmark & ---        & \checkmark & \checkmark & --- \\
$-$ Sym                                                   & \checkmark & \checkmark & \checkmark & ---        & \checkmark & --- \\
$-$ Cons                                                  & \checkmark & \checkmark & \checkmark & \checkmark & ---        & --- \\
\bottomrule
\end{tabular}
\end{table}

%% file: references.bib
@article{shen2025phyx,
  title = {{PhyX}: Does Your Model Have the ``{Wits}'' for Physical Reasoning?},
  author={Shen, Hui and Wu, Taiqiang and Han, Qi and others},
  journal={arXiv preprint arXiv:2505.15929},
  year={2025}
}

@inproceedings{he2024olympiadbench,
  title={{OlympiadBench}: A Challenging Benchmark for Promoting {AGI} with Olympiad-Level Bilingual Multimodal Scientific Problems},
  author={He, Chaoqun and Luo, Renjie and Bai, Yuzhuo and Hu, Shengding and others},
  booktitle={ACL},
  year={2024}
}

@inproceedings{xu2025ugphysics,
  title={{UGPhysics}: A Comprehensive Benchmark for Undergraduate Physics Reasoning with Large Language Models},
  author={Xu, Xin and Xu, Qiyun and Xiao, Tong and others},
  year = {2025},
  booktitle = {International Conference on Machine Learning (ICML)}
}

@article{zhang2024physreason,
  title={{PhysReason}: A Comprehensive Benchmark towards Physics-Based Reasoning},
  author={Zhang, Xinyu and Dong, Yuxuan and Wu, Yanrui and others},
  journal={arXiv preprint arXiv:2502.12054},
  year={2025}
}

@inproceedings{yue2024mmmu,
  title={{MMMU}: A Massive Multi-discipline Multimodal Understanding and Reasoning Benchmark for Expert {AGI}},
  author={Yue, Xiang and Ni, Yuansheng and Zhang, Kai and Zheng, Tianyu and others},
  booktitle={CVPR},
  year={2024}
}

@article{tsoukalas2024putnambench,
  title = {{PutnamBench}: Evaluating Neural Theorem-Provers on the {Putnam} Mathematical Competition},
  author = {Tsoukalas, George and Lee, Jasper and Jennings, John and Xin, Jimmy and others},
  journal={NeurIPS Datasets \& Benchmarks},
  year={2024}
}

@article{glazer2024frontiermath,
  title={{FrontierMath}: A Benchmark for Evaluating Advanced Mathematical Reasoning in {AI}},
  author = {Glazer, Elliot and Erdil, Ege and Besiroglu, Tamay and Chicharro, Diego and others},
  journal={arXiv preprint arXiv:2411.04872},
  year={2024}
}

@inproceedings{sainz2023nlp,
  title = {{NLP} Evaluation in trouble: On the Need to Measure {LLM} Data Contamination for each Benchmark},
  author={Sainz, Oscar and Campos, Jon Ander and Garc{\'\i}a-Ferrero, Iker and others},
  booktitle={EMNLP Findings},
  year={2023}
}

@article{singh2024beyond,
  title={Evaluation data contamination in {LLMs}: how do we measure it and (when) does it matter?},
  author={Singh, Aaditya K. and Kocyigit, Muhammed Yusuf and Poulton, Andrew and others},
  journal={arXiv preprint arXiv:2411.03923},
  year={2024}
}

@article{shao2024deepseekmath,
  title={{DeepSeekMath}: Pushing the Limits of Mathematical Reasoning in Open Language Models},
  author={Shao, Zhihong and Wang, Peiyi and Zhu, Qihao and others},
  journal={arXiv preprint arXiv:2402.03300},
  year={2024}
}

@article{yu2025dapo,
  title={{DAPO}: An Open-Source {LLM} Reinforcement Learning System at Scale},
  author={Yu, Qiying and Zhang, Zheng and Zhu, Ruofei and others},
  journal={arXiv preprint arXiv:2503.14476},
  year={2025}
}

@article{zheng2025gspo,
  title={Group Sequence Policy Optimization},
  author={Zheng, Chujie and Liu, Shixuan and Li, Mingze and others},
  journal={arXiv preprint arXiv:2507.18071},
  year={2025}
}

@article{sheng2024verl,
  title = {{HybridFlow}: A Flexible and Efficient {RLHF} Framework},
  author={Sheng, Guangming and Zhang, Chi and Ye, Zilingfeng and Wu, Xibin and others},
  journal={arXiv preprint arXiv:2409.19256},
  year={2024},
  note = {Released as the open-source \texttt{verl} framework}
}

@misc{bai2025qwen3vl,
  title={{Qwen3-VL}},
  author={{Qwen Team}},
  year={2025},
  howpublished={\url{https://huggingface.co/Qwen/Qwen3-VL-8B-Thinking}},
  note={Vision-language model release}
}

@article{chen2025internvl3,
  title = {{InternVL3}: Exploring Advanced Training and Test-Time Recipes for Open-Source Multimodal Models},
  author = {Zhu, Jinguo and Wang, Weiyun and Chen, Zhe and others},
  journal={arXiv preprint arXiv:2504.10479},
  year={2025}
}

@article{li2024llavaov,
  title={{LLaVA-OneVision}: Easy Visual Task Transfer},
  author = {Li, Bo and Zhang, Yuanhan and Guo, Dong and others},
    year={2024},
  journal = {arXiv preprint arXiv:2408.03326}
}

@article{cohen1960kappa,
  title={A Coefficient of Agreement for Nominal Scales},
  author={Cohen, Jacob},
  journal={Educational and Psychological Measurement},
  volume={20},
  number={1},
  pages={37--46},
  year={1960}
}

@article{gebru2018datasheets,
  title={Datasheets for Datasets},
  author = {Gebru, Timnit and Morgenstern, Jamie and Vecchione, Briana and Vaughan, Jennifer Wortman and Wallach, Hanna and Daum{\'e} III, Hal and Crawford, Kate},
  journal={Communications of the ACM},
  year={2021},
  volume = {64},
  number = {12},
  pages = {86--92}
}

@misc{efo,
  title = {{Estonian Physics Olympiad}: Problem Collection 2004--2018},
  author={{Estonian Physics Olympiad}},
  howpublished = {\url{https://efo.fyysika.ee/}},
  year={2018}
}

@misc{zhou2018handouts,
  title = {Physics Olympiad Handouts},
  author={Zhou, Kevin},
  howpublished={\url{https://knzhou.github.io/}},
  year = {2024}
}

@misc{ipho2025archive,
  title = {{International Physics Olympiad}: Archived Problems and Solutions},
  author = {{IPhO Unofficial Archive}},
  howpublished={\url{https://ipho-unofficial.org/}},
  year={2025},
  note = {Unofficial community-maintained archive}
}

@misc{usapho2025archive,
  title = {{U.S. Physics Olympiad}: Archived Problems and Solutions},
  author={{American Association of Physics Teachers}},
  howpublished={\url{https://www.aapt.org/physicsteam/}},
  year={2025}
}

@misc{apho2025archive,
  title = {2025 {Asian Physics Olympiad}: Problems and Solutions},
  author = {{APhO 2025 Organizing Committee}},
  howpublished = {\url{https://www.apho2025.sa/}},
  year={2025}
}

@misc{inpho2025archive,
  title = {{Indian National Physics Olympiad}: Archived Problems and Solutions},
  author={{Homi Bhabha Centre for Science Education}},
  howpublished={\url{https://olympiads.hbcse.tifr.res.in/}},
  year={2025}
}

@misc{nbpho2025archive,
  title = {{Nordic-Baltic} Physics Olympiad: Archived Problems and Solutions},
  author={{NBPhO Committee}},
  howpublished = {\url{https://nbpho.ee/}},
  year={2025}
}

@misc{eupho2025archive,
  title = {{European Physics Olympiad}: Archived Problems and Solutions},
  author={{EuPhO Committee}},
  howpublished={\url{https://eupho.ee/}},
  year={2025}
}

@misc{openstax2024physics,
  title={{College Physics 2e} and {University Physics Volumes 1-3}},
  author={{OpenStax}},
  howpublished={\url{https://openstax.org/subjects/science}},
  year={2024}
}

@misc{stackexchange2024physics,
  title = {Physics {Stack Exchange}: Question and Answer Archive},
  author={{Stack Exchange Inc.}},
  howpublished={\url{https://physics.stackexchange.com/}},
  year={2024}
}

@inproceedings{rafailov2023dpo,
  title={Direct Preference Optimization: Your Language Model is Secretly a Reward Model},
  author={Rafailov, Rafael and Sharma, Archit and Mitchell, Eric and Manning, Christopher D and Ermon, Stefano and Finn, Chelsea},
  booktitle={NeurIPS},
  year={2023}
}

@article{meng2025mmeureka,
  title = {{MM-Eureka}: Exploring the Frontiers of Multimodal Reasoning with Rule-based Reinforcement Learning},
  author = {Meng, Fanqing and Du, Lingxiao and Liu, Zongkai and Zhou, Zhixiang and Lu, Quanfeng and Fu, Daocheng and Han, Tiancheng and Shi, Botian and Wang, Wenhai and He, Junjun and Zhang, Kaipeng and Luo, Ping and Qiao, Yu and Zhang, Qiaosheng and Shao, Wenqi},
  journal={arXiv preprint arXiv:2503.07365},
  year={2025}
}

@article{guo2025deepseekr1,
  title = {{DeepSeek-R1}: Incentivizing Reasoning Capability in {LLMs} via Reinforcement Learning},
  author={DeepSeek-AI},
  journal={arXiv preprint arXiv:2501.12948},
  year={2025}
}

@inproceedings{akhtar2024croissant,
  title={Croissant: A Metadata Format for {ML}-Ready Datasets},
  author={Akhtar, Mubashara and Benjelloun, Omar and Conforti, Costanza and others},
  booktitle = {NeurIPS Datasets and Benchmarks Track},
  year={2024},
  url={https://arxiv.org/abs/2403.19546}
}

@inproceedings{qiu2025phybench,
  title = {{PHYBench}: Holistic Evaluation of Physical Perception and Reasoning in Large Language Models},
  author={Qiu, Shi and Guo, Shaoyang and Song, Zhuo-Yang and Sun, Yunbo and Cai, Zeyu and Wei, Jiashen and Luo, Tianyu and Yin, Yixuan and Zhang, Haoxu and Hu, Yi and others},
  booktitle={NeurIPS Datasets and Benchmarks Track},
  year={2025},
  note={\url{https://openreview.net/forum?id=brG8FPq1cf}}
}

@article{wang2025physunibench,
  title={{PhysUniBench}: A Multi-Modal Physics Reasoning Benchmark at Undergraduate Level},
  author={Wang, Lintao and Su, Encheng and Liu, Jiaqi and others},
  year={2025},
  eprint={2506.17667},
  archivePrefix={arXiv},
  journal = {arXiv preprint arXiv:2506.17667}
}

@article{liu2025oibench,
  title = {{OIBench}: Benchmarking Strong Reasoning Models with {Olympiad in Informatics}},
  author={Zhu, Yaoming and Wang, Junxin and Li, Yiyang and others},
    year={2025},
  journal = {arXiv preprint arXiv:2506.10481}
}

@article{phan2025hle,
  title = {{Humanity's Last Exam}},
  author={Phan, Long and Gatti, Alice and Han, Ziwen and Li, Nathaniel and Hu, Josephina and Zhang, Hugh and Zhang, Chen Bo Calvin and Shaaban, Mohamed and Ling, John and Shi, Sean and others},
  year={2025},
  eprint={2501.14249},
  archivePrefix={arXiv},
  primaryClass = {cs.LG},
  journal = {arXiv preprint arXiv:2501.14249}
}

@inproceedings{wang2024mmlupro,
  title={{MMLU-Pro}: A More Robust and Challenging Multi-Task Language Understanding Benchmark},
  author={Wang, Yubo and Ma, Xueguang and Zhang, Ge and Ni, Yuansheng and Chandra, Abhranil and Guo, Shiguang and Ren, Weiming and Arulraj, Aaran and He, Xuan and Jiang, Ziyan and others},
  booktitle={NeurIPS Datasets and Benchmarks Track},
  year={2024},
  note={Spotlight; \url{https://openreview.net/forum?id=y10DM6R2r3}}
}

@inproceedings{yue2024mmmupro,
  title={{MMMU-Pro}: A More Robust Multi-discipline Multimodal Understanding Benchmark},
  author={Yue, Xiang and Zheng, Tianyu and Ni, Yuansheng and Wang, Yubo and Zhang, Kai and Tong, Shengbang and Sun, Yuxuan and Yu, Botao and Zhang, Ge and Sun, Huan and others},
  booktitle = {ACL},
  year = {2025},
  eprint={2409.02813}
}

@inproceedings{zhang2024sciinstruct,
  title={{SciInstruct}: A Self-Reflective Instruction Annotated Dataset for Training Scientific Language Models},
  author={Zhang, Dan and Hu, Ziniu and Zhoubian, Sining and Du, Zhengxiao and Yang, Kaiyu and Wang, Zihan and Yue, Yisong and Dong, Yuxiao and Tang, Jie},
  booktitle={NeurIPS Datasets and Benchmarks Track},
  year={2024},
  note={\url{https://openreview.net/forum?id=LC1QAqhePv}}
}

@article{yang2023rephrased,
  title={Rethinking Benchmark and Contamination for Language Models with Rephrased Samples},
  author={Yang, Shuo and Chiang, Wei-Lin and Zheng, Lianmin and Gonzalez, Joseph E. and Stoica, Ion},
  year={2023},
  eprint={2311.04850},
  archivePrefix={arXiv},
  note={Demonstrates that n-gram contamination audits miss rephrased duplicates; motivates Stage-2 of our pipeline.},
  journal = {arXiv preprint arXiv:2311.04850}
}

@article{wang2024multilingualcontam,
  title={Contamination Report for Multilingual Benchmarks},
  author={Ahuja, Sanchit and Gumma, Varun and Sitaram, Sunayana},
  year={2024},
  eprint={2410.16186},
  archivePrefix={arXiv},
  note={Tests 7 LLMs on multilingual benchmarks; finds nearly all are contaminated.},
  journal = {arXiv preprint arXiv:2410.16186}
}

@article{xuan2025mmlluprox,
  title = {{MMLU-ProX}: A Multilingual Benchmark for Advanced Large Language Model Evaluation},
  author = {Xuan, Weihao and Yang, Rui and Qi, Heli and Zeng, Qingcheng and Xiao, Yunze and Feng, Aosong and Liu, Dairui and Xing, Yun and Wang, Junjue and Gao, Fan and others},
  year={2025},
  eprint={2503.10497},
  archivePrefix={arXiv},
  note={29 languages, 11{,}829 identical questions per language.},
  journal = {arXiv preprint arXiv:2503.10497}
}

@inproceedings{chow2025physbench,
  title={{PhysBench}: Benchmarking and Enhancing Vision-Language Models for Physical World Understanding},
  author={Chow, Wei and Mao, Jiageng and Li, Boyi and Seita, Daniel and Guizilini, Vitor and Wang, Yue},
  booktitle={ICLR},
  year={2025},
  note={Oral; \url{https://openreview.net/forum?id=Q6a9W6kzv5}}
}

@article{ravaut2024constam,
  title={A Comprehensive Survey of Contamination Detection Methods in Large Language Models},
  author={Ravaut, Mathieu and Ding, Bosheng and Jiao, Fangkai and others},
  year = {2025},
  eprint={2404.00699},
  archivePrefix={arXiv},
  journal = {Transactions on Machine Learning Research}
}

@inproceedings{dekoninck2024constat,
  title={{ConStat}: Performance-Based Contamination Detection in Large Language Models},
  author = {Dekoninck, Jasper and M{\"u}ller, Mark Niklas and Vechev, Martin},
  booktitle={NeurIPS},
  year={2024}
}

@article{wu2025bitter,
  title={The Bitter Lesson Learned from 2{,}000+ Multilingual Benchmarks},
  author={Wu, Minghao and Wang, Weixuan and Liu, Sinuo and others},
  year = {2025},
  eprint={2504.15521},
  archivePrefix={arXiv},
  journal = {arXiv preprint arXiv:2504.15521}
}

@article{wang2025enigmaeval,
  title={{EnigmaEval}: A Benchmark of Long Multimodal Reasoning Challenges},
  author={Wang, Clinton J. and Lee, Dean and Menghini, Cristina and others},
  year={2025},
  eprint={2502.08859},
  archivePrefix={arXiv},
  note={Frontier-model pass-rate so low that contamination is empirically dismissed.},
  journal = {arXiv preprint arXiv:2502.08859}
}

@misc{lee2024mxbai,
  title = {Open Source Strikes Bread - New Fluffy Embedding Model},
  author={Lee, Sean and Shakir, Aamir and Koenig, Darius and Lipp, Julius},
  year={2024},
  publisher={Mixedbread},
  url={https://www.mixedbread.com/blog/mxbai-embed-large-v1},
  note={\texttt{mxbai-embed-large-v1}: 1024-dim sentence embedding model used in our Stage-2 audit.}
}

@misc{anthropic2025sonnet,
  title = {{Claude Sonnet 4.5}},
  author={{Anthropic}},
  year={2025},
  url={https://www.anthropic.com/news/claude-sonnet-4-5},
  note={Model used as judge and frontier-baseline reference throughout this paper.}
}
